\documentclass{article}

\usepackage{arxiv}

\usepackage[utf8]{inputenc} % allow utf-8 input
\usepackage[T1]{fontenc}    % use 8-bit T1 fonts
\usepackage{hyperref}       % hyperlinks
\usepackage{url}            % simple URL typesetting
\usepackage{booktabs}       % professional-quality tables
\usepackage{amsfonts}  
\usepackage{float}% blackboard math symbols
\usepackage{nicefrac}       % compact symbols for 1/2, etc.
\usepackage{microtype}      % microtypography
\usepackage{lipsum}
\usepackage{array}
\usepackage{subcaption}
\usepackage{multirow}
\usepackage{natbib}
\usepackage{amsmath}
\usepackage{graphicx}
\graphicspath{ {./images/} }

\title{Leveraging Large Language Models for Bengali Math Word Problem Solving with Chain of Thought Reasoning}

\author{
 Bidyarthi Paul\\
  Department of Computer Science and Engineering\\
  Ahsanullah University of Science and Technology\\
  Tejgaon, Dhaka \\
  \texttt{bidyarthipaul01@gmail.com} \\
  \And
 Jalisha Jashim Era\\
  Department of Computer Science and Engineering\\
  Ahsanullah University of Science and Technology\\
  Tejgaon, Dhaka \\
  \texttt{ira16jalisa@gmail.com} \\
  \And
 Mirazur Rahman Zim\\
  Department of Computer Science and Engineering\\
  Ahsanullah University of Science and Technology\\
  Tejgaon, Dhaka \\
  \texttt{miraz.zim.38@gmail.com} \\
  \And
 Tahmid Sattar Aothoi\\
  Department of Computer Science and Engineering\\
  Ahsanullah University of Science and Technology\\
  Tejgaon, Dhaka \\
  \texttt{tahmidaothoi007@gmail.com} \\
  \And
 Faisal Muhammad Shah\\
  Department of Computer Science and Engineering\\
  Ahsanullah University of Science and Technology\\
  Tejgaon, Dhaka \\
  \texttt{faisal.cse@aust.edu} \\
}

\begin{document}
\maketitle
\begin{abstract}
Solving Bengali Math Word Problems (MWPs) remains a major challenge in natural language processing (NLP) due to the language’s low-resource status and the multi-step reasoning required. Existing models struggle with complex Bengali MWPs, largely because no human-annotated Bengali dataset has previously addressed this task. This gap has limited progress in Bengali mathematical reasoning. To address this, we created SOMADHAN, a dataset of 8792 complex Bengali MWPs with manually written, step-by-step solutions. We designed this dataset to support reasoning-focused evaluation and model development in a linguistically underrepresented context. Using SOMADHAN, we evaluated a range of large language models (LLMs)—including GPT-4o, GPT-3.5 Turbo, LLaMA series models, Deepseek, and Qwen—through both zero-shot and few-shot prompting with and without Chain of Thought (CoT) reasoning. CoT prompting consistently improved performance over standard prompting, especially in tasks requiring multi-step logic. LLaMA-3.3 70B achieved the highest accuracy of 88\% with few-shot CoT prompting. We also applied Low-Rank Adaptation (LoRA) to fine-tune models efficiently, enabling them to adapt to Bengali MWPs with minimal computational cost. Our work fills a critical gap in Bengali NLP by providing a high-quality reasoning dataset and a scalable framework for solving complex MWPs. We aim to advance equitable research in low-resource languages and enhance reasoning capabilities in educational and language technologies. 
\end{abstract}

% keywords can be removed

\section{Introduction}
In this era, Question Answering (QA) systems are vital in Natural Language Processing (NLP). They are designed to understand and respond to user queries in a human-like manner, crucial for applications like search engines and virtual assistants (\cite{1.1},\cite{1.2}). Improving these systems are essential due to the growing demand for quick and accurate information retrieval. The structure of QA systems varies with the domain and question types (\cite{1.6},\cite{1.7}), with Math Word Problems (MWPs) being a notable subtype.

MWPs pose significant challenges in QA systems (\cite{1.8},\cite{1.9},\cite{1.10},\cite{1.11},\cite{1.12},\cite{1.13}), requiring more than basic pattern recognition. They involve understanding mathematical operators, quantities, and their relationships to produce a solution equation. This process entails identifying numerical values, selecting appropriate operations, and forming mathematical expressions with unknown variables. Efforts to teach computers to solve MWPs date back to the 1960s, with researchers employing rule-based methods to mimic human problem-solving strategies~\cite{feigenbaum2003some}. Recent advancements in NLP and machine learning have significantly improved the ability to tackle MWPs, particularly for simpler problems.

\subsection{Chain-of-Thought Prompting}

While machine learning and transformer-based models excel at solving simpler MWPs (\cite{nayak80mathbot}), they often struggle with problems that require reasoning and multi-step solutions (\cite{3}). These MWPs demand understanding complex relationships between quantities and applying mathematical principles sequentially, which many models fail to handle effectively. Language models perform well in NLP tasks, but their reasoning abilities are limited, which cannot be solved by increasing model size (\cite{rae}). Prompting-based tools fix this with QA tasks. A study(\cite{21}) introduced CoT (Chain-of-Thought) prompting, which generates short sentences that mimic human problem-solving. 

When solving a complex reasoning task, such as a multi-step math word problem, it is natural to break the problem into smaller intermediate steps and solve each one sequentially before arriving at the final answer. For instance: ``After Era gives 2 lichies to her mom, she has 10 left. Then, after giving 3 to her dad, she has 7 left. So, the answer is 7.” A chain of thought is a series of intermediate natural language reasoning steps that lead to the final output, and we refer to this approach as chain-of-thought prompting (\cite{21}). Recent advancements on chain-of-thought prompting, have shown promise in enhancing the reasoning capabilities of large language models for solving complex MWPs. Chain-of-thought prompting involves breaking down a problem into intermediate reasoning steps, which are then solved sequentially to reach the final answer.

\begin{figure}[h]
  \includegraphics[width=\columnwidth]{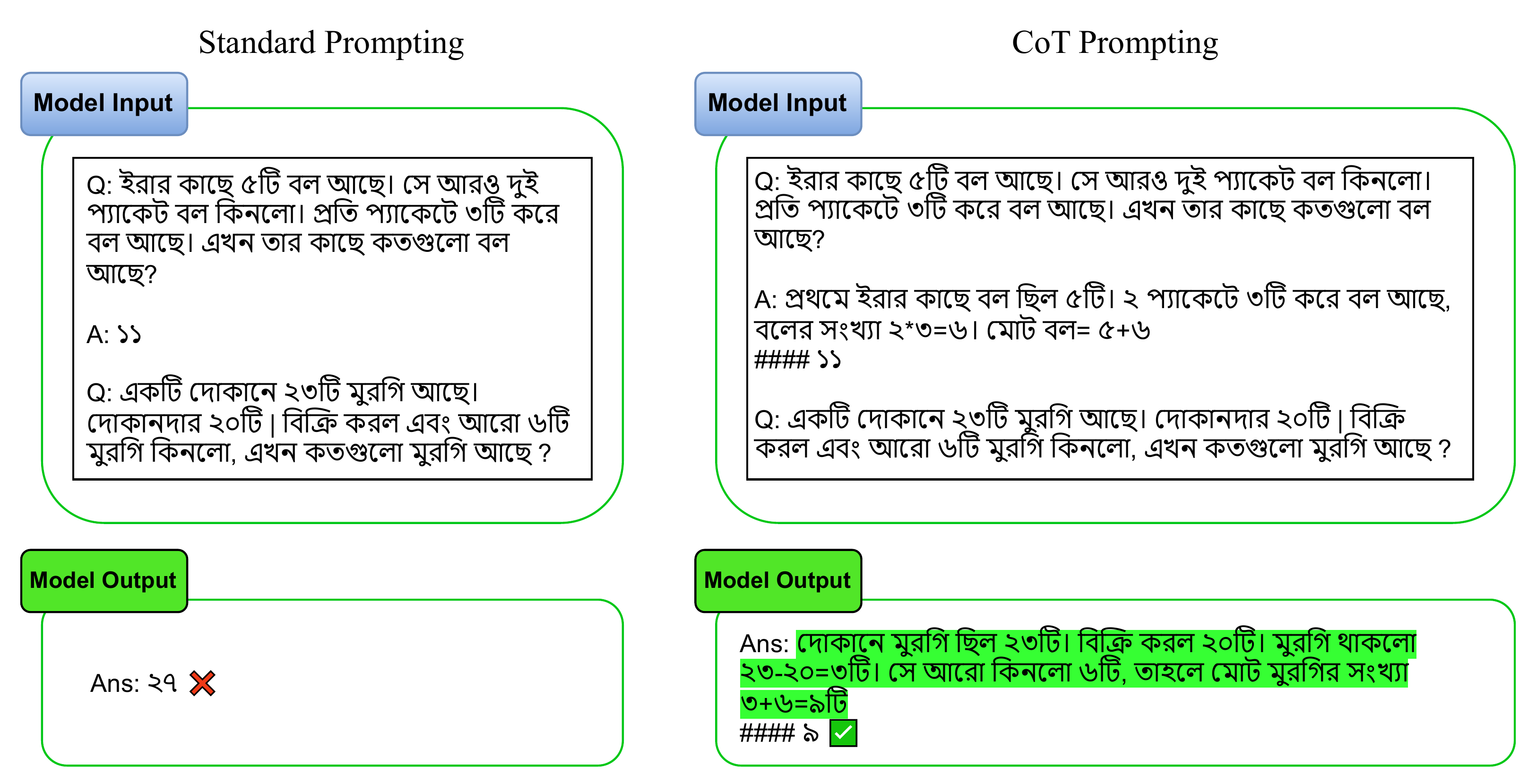}
  \caption{Chain-of-Thought (CoT) Prompting enables LLMs (Large Language Model)s to improve complex reasoning of MWPs (Math Word Problems)}
  \label{fig:introfig}
\end{figure}

 A comparison between Standard Prompting and Chain-of-Thought (CoT) Prompting for solving math word problems (MWPs) using a language model is shown in Figure~\ref{fig:introfig}. In Standard Prompting, the model is presented with the question directly, without any reasoning steps, leading to incorrect answers due to a lack of logical breakdown. In contrast, CoT Prompting provides a step-by-step explanation, allowing the model to reason through the problem logically and arrive at the correct solution. By breaking the problem into smaller parts and solving it systematically, CoT significantly improves the model's ability to handle complex reasoning tasks, as demonstrated by the accurate outputs in the CoT example.

 Studies (\cite{21}) have demonstrated that incorporating chain-of-thought reasoning into few-shot prompting significantly improves model performance on tasks such as math word problems, commonsense reasoning, and symbolic manipulation. By leveraging this method, large language models can better handle the intricate reasoning required for complex MWPs.

The challenge is even greater for low-resource languages like Bengali, where limited datasets and linguistic complexities further hinder the development of reasoning-based solutions. Bengali is a low-resource language for Math Word Problem (MWP) solving using Large Language Models, with 272.7 million speakers and seventh worldwide(\cite{wiki}). Bengali language processing tools and models are difficult to develop due to a lack of datasets, linguistic resources, standardization, and dialectal variations, which hinders LLMs for MWPs solving.

In this study, we introduce a new dataset, SOMADHAN (Solution), which focuses on complex Bengali math word problems and includes intermediate reasoning steps (chain of thought) for each solution. We create this dataset to address the lack of high-quality, human-annotated resources for Bengali MWPs, a gap that severely limits the development and evaluation of reasoning-capable language models in low-resource languages. Unlike existing datasets, which are either automatically translated or limited to final-answer supervision, SOMADHAN provides detailed, step-by-step logical reasoning, enabling models to learn and demonstrate multi-step problem-solving—an essential skill for real-world mathematical understanding. Additionally, we utilized the publicly available "PatiGonit" dataset from a Github repository, which contains equation-based simple MWPs, to analyze models' ability to handle simple MWPs. These two datasets highlight the limitations of large language models, which can only solve straightforward math problems but struggle with complex ones requiring reasoning. Our proposed pipeline for Chain-of-Thought prompting, built on the SOMADHAN dataset, addresses this challenge by enabling large language models to tackle more reasoning-intensive tasks. Several large language models, including GPT-4o\citep{gpt4o}, GPT-3.5 Turbo-16k\citep{gpt3.5}, Llama-3 8B and 70B\citep{Llama3}, llama-3.1-70b-versatile and 8b\citep{Llama3},llama-3.2-1b-preview and llama-3.2-3b-preview\citep{Llama3}, llama-3.2-90b-text-preview\citep{Llama3}, llama-3.3-70b\citep{Llama3}, deepseek-r1-distill-qwen-32b,\citep{Deepseek}, deepseek-r1-distill-llama-70b\citep{Deepseek}, qwen-2.5-32b, and qwen-qwq-32b\citep{Qwen} were evaluated for their ability to generate intermediate steps and equations. Chain-of-thought (CoT) prompting consistently outperformed standard prompting. Among all the models evaluated, Llama-3.3 with CoT and few-shot prompting achieved the highest accuracy of 88\%, outperforming all other models in both zero-shot and few-shot settings. GPT-4o showed strong performance in COT with few-shot learning across both datasets. For further experiments, we introduced LoRA (Low-Rank Adaptation) fine-tuning. LoRA was employed to adapt the models efficiently by modifying only a small subset of parameters, making it computationally and memory efficient. The goal of using LoRA was to see how well it could enhance the performance of large models like Llama-3 (8B) for the specific structure of Bengali math word problems while retaining their versatility for other general tasks.

The following are the main contributions of our study:
\begin{itemize}
    \item Developed SOMADHAN dataset containing 8,792 complex Bengali Math Word Problems with their corresponding step by step solutions.
    \item Introduced a pipeline for applying chain-of-thought prompting using the SOMADHAN dataset, enabling large language models to handle complex reasoning tasks effectively.
    \item Introduced two Prompts to solve Complex Bengali Math Word Problems that requires reasoning steps. 
    \item Performed a comparative analysis of GPT-4o, GPT-3.5 Turbo-16k, Llama-3 8B and 70B, llama-3.1-70b-versatile, llama-3.1-8b, llama-3.2-1b-preview, llama-3.2-3b-preview, llama-3.2-90b-text-preview, llama-3.3-70b, deepseek-r1-distill-qwen-32b, deepseek-r1-distill-llama-70b, qwen-2.5-32b, and qwen-qwq-32b demonstrating the strengths of zero-shot prompting, few-shot prompting and fine-tuning to solve MWPs.
    \item Employed Low-Rank Adaptation (LoRA) to efficiently fine-tune large language models for Bengali math word problems, optimizing performance with minimal computational overhead.

\end{itemize}

\section{Related Works}
In this section, we discuss works related to our study. During our exploration, we found that no prior work has focused on solving Bengali Math Word Problems (MWPs) with Chain of Thoughts reasoning. However, there have been several studies on solving MWPs in English, which we categorize into six main approaches: Traditional Deep Learning Approaches, Transformer-Based Approaches, Hybrid and Ensemble Models, Methods utilizing intermediate steps, Prompting-based approaches, and Chain-of-Thought (CoT) Reasoning Approaches. Each of these categories is discusses in the following sections.

\subsection{Traditional Deep Learning Approaches} 

These methods are increasingly being used to reduce manual effort and to enhance the performance of MWP solvers. One recent study~\cite{my_iccit} proposed a Bengali Word Problem dataset and set a Benchmark Evaluation for it using transformer and different Neural Machine Translation Models. 97.30\% accuracy was achieved via mT5. Another notable method involves using Sequence-to-Sequence (Seq2Seq) models to improve MWP solvers. A study by \cite{8} proposed a Seq2Seq RNN model to solve math word problems (MWPs), integrating a GRU-based encoder, LSTM-based decoder, and a similarity-based retrieval mechanism for improved accuracy. While the model achieved 58.1\% accuracy on Math23k and 16.1\% on Alg514, the hybrid approach boosted performance to 64.7\% and 70.1\%, respectively. However, it struggled with diverse datasets and failed to generate novel equation templates. To address this, another study by \cite{9} introduced equation normalization and explored Seq2Seq architectures, including BiLSTM, ConvS2S, and Transformer models. Their ensemble approach achieved 68.4\% accuracy on Math23k, outperforming individual models (66.7\% for BiLSTM, 64.2\% for ConvS2S, and 62.3\% for Transformer), emphasizing normalization and model diversity. In related work, \cite{10} developed the Variational Neural Machine Translation (VNMT) model with latent variables to enhance semantic understanding, achieving BLEU scores of 32.07 (Chinese-English) and 19.58 (English-German), outperforming traditional models but facing challenges with long sentences.

An improvement to Seq2Seq models was proposed by \cite{11}, who incorporated Copy and Alignment mechanisms, further optimized via Reinforcement Learning (RL), achieving accuracies of 44.5\% on Alg514, 64.0\% on NumWord, and 23.3\% on Dolphin18KT6. Their hybrid model significantly boosted performance to 82.5\%, 65.8\%, and 33.2\%, respectively. In another approach, \cite{12} introduced the Graph-to-Tree (Graph2Tree) model, using graph-based encoders and tree-based decoders to effectively capture numerical relationships. It achieved accuracies of 83.7\% on MAWPS and 77.4\% on Math23k, addressing prior gaps in quantity reasoning. Similarly, \cite{13} utilized BiGraphSAGE encoders and tree-based decoders in their Graph2Tree model, achieving 78.8\% on MAWPS and 69.65\% on MathQA but faced challenges in fully incorporating structural reasoning.

To further enhance MWP solvers, \cite{14} proposed a multi-encoder, multi-decoder framework combining sequence-based and graph-based encoders with sequence and tree decoders. This approach improved equation generation and achieved 78.4\% on the Math23k test set but still had gaps in leveraging textual structure. In another advancement, \cite{15} introduced MWP-BERT, a numeracy-augmented pre-trained language model tailored for MWPs, achieving state-of-the-art results of 84.7\% on Math23k, 76.2\% on MathQA, and 91.2\% on Ape-clean. Separately, \cite{16} presented BERT, a bidirectional pre-trained model for language understanding, achieving 93.2\% on SQuAD v1.1 and 83.1\% on SQuAD v2.0. Building upon BERT, \cite{17} enhanced it with RoBERTa by removing Next Sentence Prediction (NSP), applying dynamic masking, and using larger datasets, achieving state-of-the-art results like 90.2\% on MNLI and 96.4% on SST-2.

In a generative learning context, \cite{18} introduced GPT-2, a generative pre-trained transformer excelling in zero-shot learning tasks, with notable performance on datasets such as LAMBADA (63.2\%). For rule-based reasoning, \cite{22} presented RuleTakers for reasoning over natural language rules, achieving 99\% on synthetic datasets and over 90\% on hand-authored rules, though paraphrased reasoning remained a challenge. Extending this work, \cite{23} proposed RuleBERT, fine-tuning RoBERTa for probabilistic logic reasoning, achieving up to 99\% accuracy on overlapping rules. Finally, \cite{24} introduced EVR, a T5-based model for explainable reasoning, excelling in multi-hop tasks with interpretable steps, achieving 97.0\% on DU5 and up to 98.1\% on Birds-Electricity tasks. These advancements demonstrate significant progress in NLP for MWPs, reasoning, and language understanding across diverse datasets.

\subsection{Transformer-Based Approaches}

In a recent study, \cite{trans1} revisited MWP-BERT, highlighting its use of numeracy-augmented pre-training for math word problem solving. By integrating numerical reasoning and symbolic placeholders, the model addressed gaps in capturing numerical variations in context. MWP-BERT achieved accuracies of 84.7\% on Math23k, 76.2\% on MathQA, and 91.2\% on Ape-clean. The datasets used included Math23k, MathQA, and Ape-clean (81,225 filtered problems from Ape210K). In another contribution, \cite{trans2} introduced a novel approach for improving MWP solvers by generating linguistic variants of problem statements using GPT-3. These paraphrased problems were solved using a DeBERTa-based solver, with majority voting employed for final predictions. This approach addressed the lack of robustness in existing models to paraphrased problems. The method achieved accuracies of 91.0\% on MAWPS, 79.1\% on PARAMAWPS, and 63.5\% on SVAMP. The datasets included MAWPS, PARAMAWPS, and SVAMP.

% \cite{trans1} revisited MWP-BERT, highlighting its use of numeracy-augmented pre-training for math word problem solving. By integrating numerical reasoning and symbolic placeholders, the model addressed gaps in capturing numerical variations in context. MWP-BERT achieved accuracies of 84.7\% on Math23k, 76.2\% on MathQA, and 91.2\% on Ape-clean. The datasets used included Math23k, MathQA, and Ape-clean (81,225 filtered problems from Ape210K). \cite{trans2} introduced a novel approach for improving MWP solvers by generating linguistic variants of problem statements using GPT-3. These paraphrased problems were solved using a DeBERTa-based solver, with majority voting employed for final predictions. This approach addressed the lack of robustness in existing models to paraphrased problems. The method achieved accuracies of 91.0\% on MAWPS, 79.1\% on PARAMAWPS, and 63.5\% on SVAMP. The datasets included MAWPS, PARAMAWPS, and SVAMP.

\subsection{Hybrid and Ensemble Models}
Hybrid models that combine different approaches have shown significant improvements. An ensemble model by \cite{trans3} utilizing BiLSTM and LSTM with equation normalization achieved 69.2\% accuracy on the Math23K dataset . In the paper \cite{trans4} the GTS model, integrating GRU, TreeDecoder, and gated feedforward networks, achieved 74.3\% on Math23K and 83.5\% on the MAWPS single operation dataset. WARM has been introduced in \cite{trans5} and it uses a weakly supervised approach with a bidirectional GRU encoder and three fully connected networks as the decoder, achieving 66.9\% on All Arith and 56.0\% on Math23K .

\begin{table*}[h]
\caption{A Summary of Existing Works in Solving Math Word Problems}
\label{table:tabrel}
\normalsize
\centering
\begin{tabular}{|>{\raggedright\arraybackslash}p{2.8cm}|>{\raggedright\arraybackslash}p{2.3cm}|>{\centering\arraybackslash}p{2.5cm}|>{\raggedright\arraybackslash}p{1.9cm}|>{\raggedright\arraybackslash}p{1.9cm}|>{\centering\arraybackslash}p{1.5cm}|}
\hline
\textbf{Paper Title} & \textbf{Approach} & \textbf{Model} & \textbf{Dataset} & \textbf{Accuracy} & \textbf{Language} \\ \hline

Empowering Bengali Education with AI: Solving Bengali Math Word Problems through Transformer Models \cite{my_iccit} & Transformer Based Approach, Neural Machine Translation (NMT) & Transformer, mT5, BanglaT5, mBART50 & PatiGonit & PatiGonit: 97.30\% & Bengali \\ \hline

Math word problem solving by generating linguistic variants of problem statements \cite{trans2} & Enhanced Mask Decoder and a Voting Mechanism & DeBERTa & MAWPS, PARAMAWPS & MAWPS: 91.0\%, PARAMAWPS: 79.1\% & English \\ \hline

% Graph-to-tree learning for solving math word problems \cite{12} & Graph2Tree & Graph Transformer with GCNs, Tree-based Decoder (pre-order traversal) & MAWPS, Math23K & MAWPS: 83.7\%, Math23K: 77.4\% & English \\ \hline

Chain-of-thought prompting elicits reasoning in large language models \cite{21} & Chain-of-Thought Prompting & GPT-3, LaMDA, PaLM, UL2, Codex & GSM8K & GSM8K: 63.1\% & English \\ \hline

Making Large Language Models Better Reasoners with Step-Aware Verifier \cite{19} & Diverse Verifier on Reasoning Step & OpenAI Models: davinci, text-davinci-002, code-davinci-002 & GSM8K & GSM8K: 83.2\% & English \\ \hline

Self-consistency improves chain of thought reasoning in language models \cite{20} & Self-Consistency, Chain-of-Thought Prompting (CoT) & GPT-3: code-davinci-001, code-davinci-002, LaMDA-137B, PaLM-540B & GSM8K & GSM8K: 83.2\% & English \\ \hline
\end{tabular}
\end{table*}

\subsection{Methods Utilizing Intermediate Steps} 
Numerous studies have demonstrated the many benefits of teaching neural networks to produce intermediate steps through fine-tuning or training. Natural language intermediate steps, in particular, have shown promise in enhancing model robustness and interpretability. For instance, \cite{29} introduced the "Rationalize-Then-Predict" framework, a two-stage method aimed at improving model robustness in adversarial contexts by using rationalizers to extract relevant inputs before prediction. Although models like VIB and SPECTRA showed promise (e.g., 82.6\% accuracy on FEVER), they remained vulnerable to strong attacks. In a related effort, \cite{30} proposed BabbleLabble, a weak supervision approach leveraging natural language explanations and semantic parsers, achieving an F1 score of 50.1 on the Spouse dataset but facing challenges with noisy labels and generalization. Similarly, \cite{31} highlighted the utility of intermediate annotations in reading comprehension, reporting 85\% accuracy with Random Forest but acknowledging limited dataset diversity. Another study by \cite{25} explored annotator rationales, improving SVM performance to 92.2\% on the Polarity Dataset but identified issues with annotation quality and domain applicability. To address explanation integration, \cite{26} developed REMOTE, which incorporated human-provided explanations to enhance language models, achieving 62.0\% on HatEval and 92.7\% on AmazonMusic, but faced scalability issues with labeled data. Further, \cite{27} evaluated the utility of explanations for improving task performance, achieving 91.41\% on e-SNLI while noting inconsistent benefits across tasks. Additionally, \cite{28} introduced DREAM, which refined internal representations in QA models like Macaw, improving accuracy by 4\% on CODAH but exposing limitations in coherent scene modeling. In a related context, \cite{32} proposed Learning with Latent Language (L3) to parameterize multitask scenarios, showing promise in structured tasks like ShapeWorld (70\% accuracy) but struggling with generalizing to abstract concepts. These studies collectively emphasize advancements in integrating explanations and rationales into AI systems while highlighting persistent challenges in scalability, robustness, and generalization.

In support of explanation-based modeling, \cite{33} introduced the e-SNLI dataset, a natural language inference (NLI) benchmark that incorporates explanations, enabling models to predict decisions and justify them. Models like e-INFERSENT achieved 83.96\% accuracy, while EXPLAIN THE PREDICT ATTENTION scored 81.71\%, though reliance on spurious correlations in existing models revealed gaps in robustness. Moreover, \cite{34} developed the CoS-E dataset and CAGE framework for commonsense reasoning, which improved model accuracy to 72.6\% on the CoS-E dev split but highlighted limitations in explanation generation and dataset availability, underscoring the importance of leveraging annotated explanations for better interpretability. Finally, \cite{36} proposed the Self-Taught Reasoner (STaR), which combines reasoning and rationalization with models like GPT-J and BERT, achieving notable accuracies, such as 72.6\% on CommonsenseQA and GSM8K. However, STaR faced challenges in scalability, requiring significant resources for rationale generation and balancing the trade-off between accuracy and explainability. Collectively, these works highlight the potential of explanation-based approaches in enhancing transparency, reasoning, and trust in AI systems while identifying gaps in dataset availability and model generalization.

\subsection{Prompting-Based Approaches} 

Recent advancements in prompting techniques have led to significant improvements in the performance of language models. Building on the concept of few-shot prompting introduced by \cite{2}, various methods have been explored to refine and optimize how models interact with input data. For example, \cite{37} proposed automatic prompt learning techniques, enabling models to adapt their prompts based on the task at hand, rather than relying solely on predefined prompts. Additionally, studies such as those by \cite{38}, \cite{39}, and \cite{40} have focused on providing models with task-specific instructions, allowing for better alignment with the desired output.

Another significant development in this area is the use of detailed task instructions, which has been shown to further enhance the performance of language models. In particular, \cite{41}, along with works by \cite{38}, \cite{40}, \cite{39}, and \cite{42}, emphasize that enriching input-output pairs with clear and specific guidance helps improve model understanding and reasoning, especially in complex tasks like math word problems. This trend of augmenting prompts with additional context and instructions has been pivotal in achieving state-of-the-art results in various natural language processing tasks, demonstrating the importance of prompt design in optimizing model behavior.

% Recent advancements in prompting techniques have led to significant improvements in the performance of language models. Building on the concept of few-shot prompting introduced by \cite{2}, various methods have been explored to refine and optimize how models interact with input data. For instance, \cite{37} proposed automatic prompt learning techniques, enabling models to adapt their prompts based on the task at hand, rather than relying solely on predefined prompts. In addition, recent works such as \cite{38}, \cite{39}, and \cite{40} have focused on providing models with task-specific instructions, allowing for better alignment with the desired output.

% Another significant development in this area is the use of detailed task instructions, which has been shown to further enhance the performance of language models. \cite{41}, \cite{38}, \cite{40}, \cite{39}, and \cite{42} emphasize that enriching input-output pairs with clear and specific guidance helps improve model understanding and reasoning, especially in complex tasks like math word problems. This trend of augmenting prompts with additional context and instructions has been pivotal in achieving state-of-the-art results in various natural language processing tasks, demonstrating the importance of prompt design in optimizing model behavior.

\subsection{Chain-of-Thought (CoT) Reasoning Approaches} 

Chain of Thought (CoT) prompting has become a widely adopted method for improving the performance of language models, especially in tasks requiring multi-step reasoning. By breaking down a complex problem into smaller, more manageable steps, CoT helps models generate clearer and more accurate answers. For example, \cite{21} demonstrated that CoT prompting significantly outperforms traditional methods for solving math word problems (MWPs), enhancing both the accuracy of the results and the transparency of the model’s reasoning process.

Building on this foundation, \cite{19} introduced DIVERSE (Diverse Verifier on Reasoning Step), a framework designed to improve language model reasoning by generating diverse prompts. This approach explores multiple reasoning paths for the same question and uses a verifier to filter out incorrect answers through a weighted voting system. Importantly, DIVERSE focuses on evaluating each reasoning step individually rather than the entire chain, which further refines the model's decision-making process. In a related effort, \cite{20} introduced the concept of self-consistency, a decoding strategy that selects the most consistent answer from multiple reasoning paths, providing another layer of validation. When applied to arithmetic reasoning benchmarks, self-consistency has been shown to enhance CoT prompting, yielding more accurate and reliable results.

Our study explores methods for solving Math Word Problems (MWPs), categorized into Deep Learning, Transformer-based, Hybrid, Prompting, and Chain-of-Thought (CoT) Reasoning approaches. The most significant gap identified in this study is the complete absence of prior work addressing Math Word Problems (MWPs) in the Bengali language. This highlights a critical underrepresentation of Bengali in existing research on mathematical reasoning and natural language processing. In contrast, for English datasets, traditional models like Seq2Seq and Graph2Tree advanced MWP solving but struggled with diverse datasets and reasoning structures. Transformer-based models, including MWP-BERT and GPT-3, achieved state-of-the-art results but faced robustness issues. Prompting and CoT methods improved multi-step reasoning by leveraging task-specific instructions and diverse reasoning paths. A summary of the existing works is given in Table~\ref{table:tabrel}.

\section{Problem Description}
The problem is best framed as a model's ability to solve complex math word problems by generating intermediate reasoning steps and a final result. Evaluation focuses solely on the correctness of the final result, disregarding intermediate steps.

Let, x denotes the input math word problem, f(x) denotes the output generated by the model, consisting of a sequence of intermediate reasoning steps followed by a final result $r_{pred}$, and $r_{true}$ denotes the ground truth final result for the problem.

The model's output can be represented as:
\begin{center}
    f(x) = ($s_{1}$, $s_{2}$, ...... ,$s_{k}$, $r_{pred}$)
\end{center}
where $s_{1}$, $s_{2}$, ...... ,$s_{k}$ are the intermediate steps and $r_{pred}$ is the predicted final result.

% For evaluation, the intermediate reasoning steps $s_{1}$, $s_{2}$, ...... ,$s_{k}$ are not considered. Only the final result $r_{pred}$ is compared with the ground truth $r_{true}$:

% \begin{center}
%     \text{Evaluation}(f(x), $r_{\text{true}}$) = 
%     \begin{equation}
%     1 & \text{if } $r_{\text{pred}} = r_{\text{true}}$ \\
%     0 & \text{if } $r_{\text{pred}} \neq r_{\text{true}}$
%     \end{equation}
% \end{center}

% The model is considered to have successfully solved the problem if: $r_{\text{pred}} = r_{\text{true}}$

% If the dataset contains \textit{N} problems {$x_{1}$, $x_{2}$, ...... , $x_{\textit{N}}$}, the accuracy \textit{A} is computed as:
% \begin{center}
%    A = $\frac{1}{N}$ $\sum_{i=1}^{N}$ $\mathbf{1}(r_{\text{pred}, i} = r_{\text{true}, i})$ 
% \end{center}

% where $\mathbf{1}(\cdot)$ is the indicator function, which equals 1 if the condition is true and 0 otherwise.

\section{Corpus Creation}
\begin{figure}[h]
  \includegraphics[width=\columnwidth]{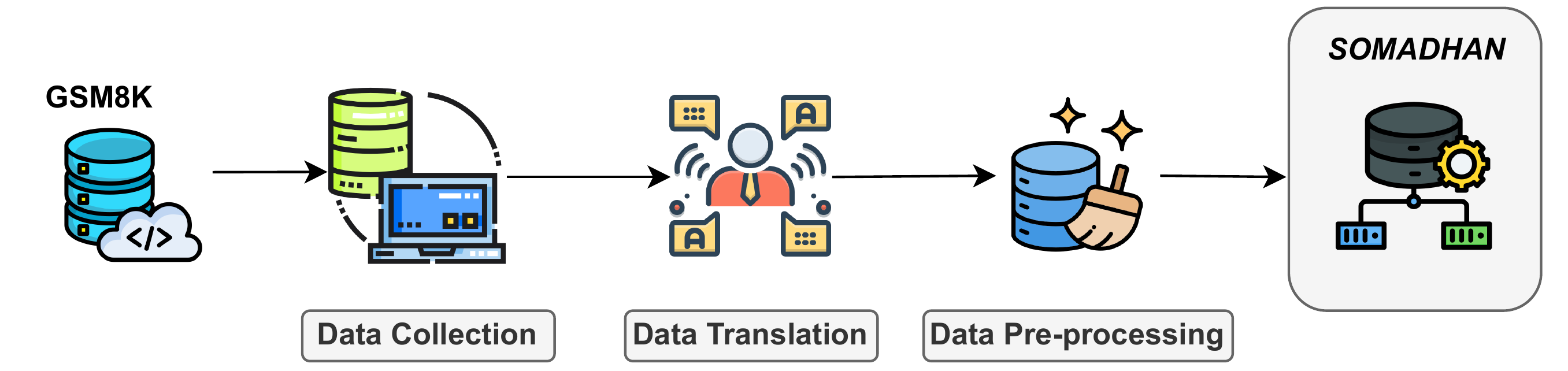}
  \caption{Pipeline for the development of the SOMADHAN dataset}
  \label{fig:datafig}
\end{figure}

As per our exploration, we found no publicly available dataset addresses complex Bengali Math Word Problems (MWPs) with detailed reasoning steps and human-verified solutions. While a few Bengali MWP datasets exist, they primarily rely on automated translation, which leads to poor linguistic quality and a lack of meaningful reasoning structure. These limitations make them unsuitable for training or evaluating models that aim to perform multi-step mathematical reasoning in Bengali. To address this gap, we developed a new Bengali math word problem dataset, SOMADHAN. This dataset has been meticulously manually annotated by expert annotators, ensuring higher accuracy and contextual relevance in solving complex math problems with reasoning steps and solutions. This dataset focuses on problems that require advanced reasoning and step-by-step solutions, designed to facilitate the evaluation of large language models' reasoning capabilities. Figure~\ref{fig:datafig} illustrates the pipeline we employed to develop the SOMADHAN dataset. 

\subsection{Data Collection}
\textbf{SOMADHAN Dataset:} To develop our Bengali math word problem dataset with intermediate steps and answers, we required a substantial collection of problems reflecting complex reasoning processes. As a foundation, we utilized the publicly available GSM8K (Grade School Math)\footnote{\url{https://github.com/openai/grade-school-math/tree/master/grade_school_math/data}} dataset~\citep{3}, which contains a diverse set of English grade school math problems. GSM8k has four parts: train, train socratic, test, and test socratic. But specifically, we leveraged the train and test portions, comprising a total of 8,792 questions and solutions. GSM8K was selected for its comprehensive coverage of grade school math concepts, making it an ideal base for developing our dataset. Inspired by GSM8K, we created SOMADHAN (Bengali Grade School Math). Our dataset comprises of 8,792 Bengali Math Word Problems along with their step by step solutions.

\textbf{PatiGonit Dataset:} The second dataset used in this study, PatiGonit, was publicly available and collected from a \href{https://github.com/Bidyarthi404/PatiGonit}{Github} repository. It consists of 10,000 simple Bengali math word problems, each paired with a corresponding equation as the solution. The dataset primarily focuses on elementary school-level problems, encompassing basic arithmetic and some introductory algebraic challenges. The problems are organized into two columns: one for the word problem and the other for the corresponding mathematical equation. PatiGonit serves as a valuable resource for analyzing simple equation-based math word problems.

\subsection{Data Translation}
In this section, we describe the dataset translation process for the SOMADHAN dataset. Manual data translation is inherently challenging as it requires not only linguistic accuracy but also the preservation of contextual meaning, particularly for reasoning-based tasks in math word problems. Any loss of reasoning structure during translation could compromise the dataset's usability for downstream applications like reasoning evaluation and problem-solving.  

To ensure the reasoning quality and contextual fidelity of the translated texts, we employed expert human translators for the manual translation process. Comprehensive guidelines were established to provide translators with clear instructions on preserving mathematical and contextual details. These predefined standards, combined with the translators' expertise and judgment, ensured the integrity and consistency of the translated dataset.

\subsubsection{Translator Indentity}
For the translation process, we engaged professionals with expertise in English-to-Bengali translation to ensure that the translations were accurate, contextually consistent, and aligned with the reasoning requirements of the original texts. A team of five skilled individuals collaboratively worked on the translation task. The entire corpus of the SOMADHAN dataset was evenly distributed among them to ensure efficiency and maintain uniform quality across the translated content. The information regarding their expertise, experiments and other details are presented in Table~\ref{tab:infotable}.

\begin{table}[htbp]
\centering
\caption{Detailed Informations of Translators}
\begin{tabular}{ c c c c c c }
\hline
Details & Translator 1 & Translator 2 & Translator 3 & Translator 4 & Translator 5 \\ \hline
Role & Graduate & Under-graduate & Under-graduate & Under-graduate & Under-graduate \\
Age & 30 & 23 & 24 & 24 & 22 \\ 
Research field & NLP & NLP & NLP & NLP & NLP \\ 
Experience & 4 years & 2 years & 1 years & 2 years & 1 years \\ \hline
\end{tabular}

\label{tab:infotable}
\end{table}

\begin{figure}[h]
\centering
  \includegraphics[width=0.9\columnwidth]{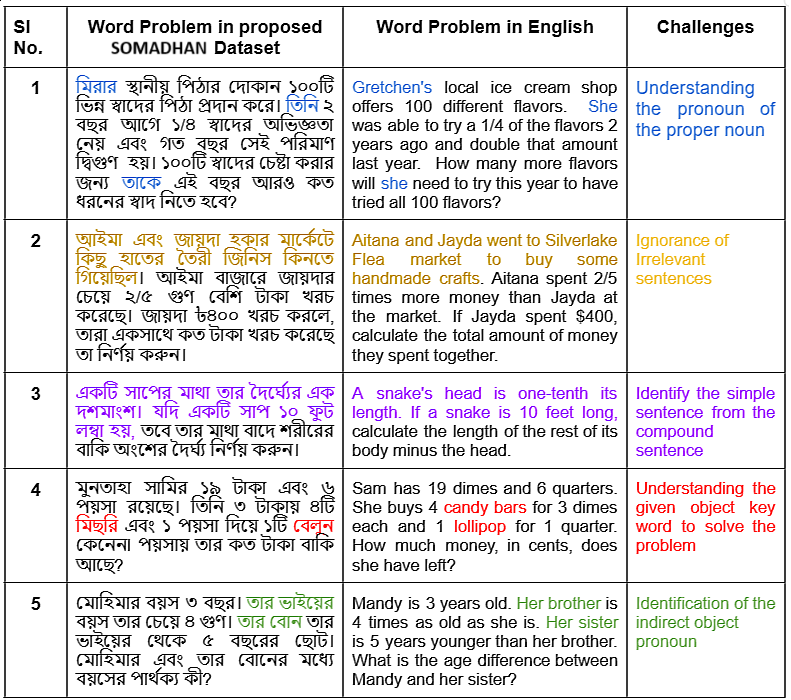}
  \caption{Challenges faced during the translation of BGSM8K dataset}
  \label{fig:probfig}
\end{figure}

\subsubsection{Translation guidelines}
To ensure accuracy, cultural relevance, and linguistic diversity, the following guidelines were established and provided to the translators:

\begin{itemize}
    \item Translations should retain the original meaning and reasoning of the text, ensuring that mathematical logic and contextual nuances are preserved.
    \item The Dollar symbol was consistently replaced with the Bengali Taka symbol to align with regional monetary conventions.
    \item English numerals were replaced with their Bengali counterparts to maintain consistency with the Bengali script.
    \item Individual names were replaced with culturally appropriate Bengali names.
    \item Similarly, Location names, objects, and food names were substituted with Bengali equivalents to ensure cultural relevance while maintaining the problem's original structure and meaning.
\end{itemize}

\subsubsection{Challenges During Translation}
Even with the predefined guidelines, translators encountered several challenges during the translation process, which are summarized in Figure~\ref{fig:probfig}.

\subsection{Dataset Statistics}

\begin{figure}[h]
\centering
  \includegraphics[width=1.0\columnwidth]{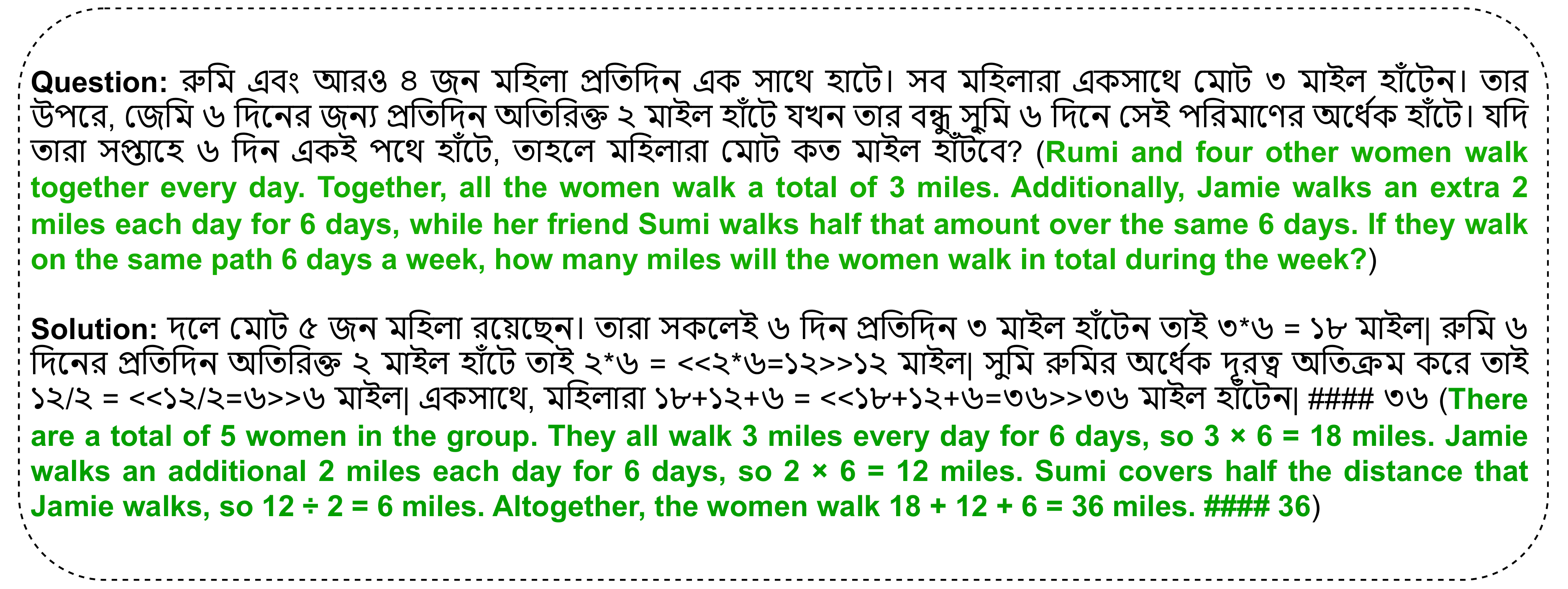}
  \caption{Sample example of SOMADHAN Dataset}
  \label{fig:sam3}
\end{figure}

The SOMADHAN dataset is composed of 8,792 Bengali math word problems with intermediate reasoning steps and their corresponding answers. Currently, the dataset includes 4,000 manually annotated samples, each containing step-by-step reasoning and a final answer. Annotation of the remaining problems is ongoing, and we plan to release the complete version in a future update. Once finalized, we will publish Version 2 of the dataset on Mendeley Data to ensure continued accessibility and reproducibility. A sample is shown in Figure~\ref{fig:sam3}.

\begin{figure}[h]
\centering
\begin{minipage}{1.0\textwidth}
  \centering
\includegraphics[width=1.0\textwidth]{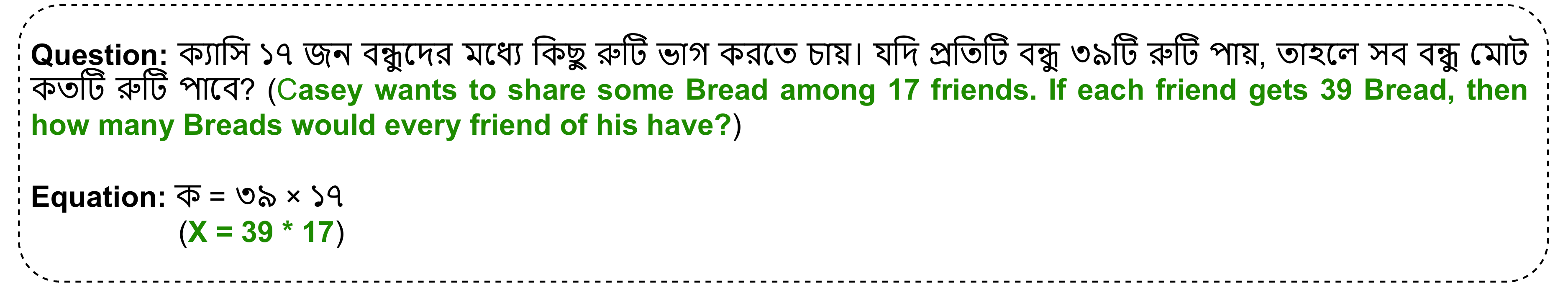}
\subcaption{Simple Equation}\label{fig:sam1}
\end{minipage}%

\begin{minipage}{1.0\textwidth}
  \centering
\includegraphics[width=1.0\textwidth]{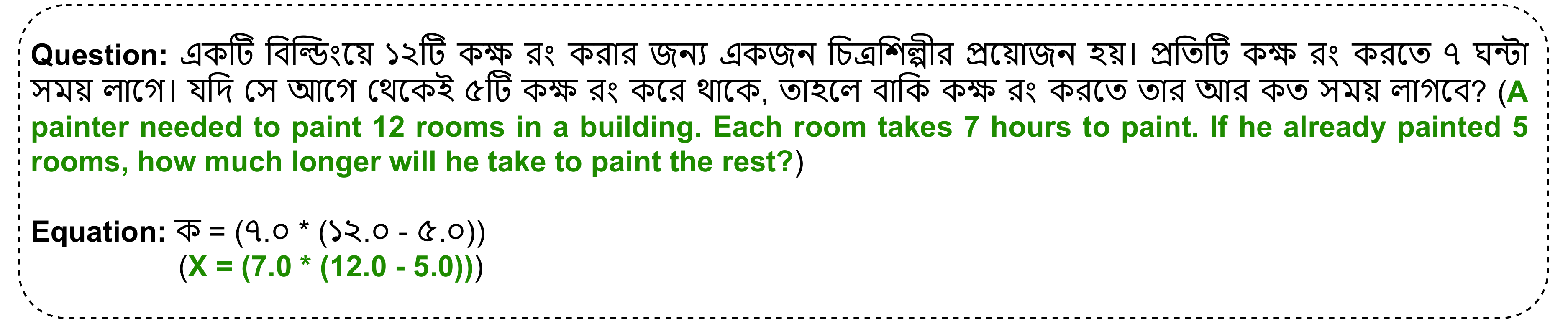}
\subcaption{Complex Equation}\label{fig:sam2}
\end{minipage}%

\caption{Sample example of PatiGonit dataset} \label{fig:sam}
\end{figure}

The PatiGonit dataset on the other hand comprises 10,000 equation-based Bengali math word problems. The dataset contains a variation of simple and complex equations; simple equations are defined as those containing only one mathematical operation, while complex equations involve multiple operations. Sample of the dataset is shown in Figure~\ref{fig:sam}.

\subsection{Data Availability}

The SOMADHAN dataset, which contains Complex Bengali Grade School Math Word Problems, is publicly available for research and academic purposes. Researchers interested in utilizing the dataset can access it through the following link:\href{https://data.mendeley.com/datasets/34bs5cxk9j/1}{Dataset of SOMADHAN (Original data)} (Mendeley Data)

\section{Methodology}
In this study, our methodology centers on the use of large language models (LLMs) for solving Bengali math word problems. Specifically, we examined the performance of LLMs by constructing tailored prompts, which were then input into the models via API configurations. The outputs were thoroughly analyzed to assess the effectiveness of the models in addressing the challenges posed by Bengali math word problems. This approach allows us to highlight the comparative advantages of using LLMs for these types of tasks. The study utilizes two distinct datasets to evaluate model performance. The primary dataset, SOMADHAN, was specifically designed for Bengali math word problems and incorporated Chain of Thought (CoT) prompting to guide the models through a logical, step-by-step reasoning process. This setup was intended to assess the ability of LLMs to solve more complex problems requiring multi-step reasoning. In contrast, the PatiGonit dataset was employed to investigate the performance of LLMs in solving simpler equation-based problems in compared to Reasoning-based problems.

\subsection{Evaluation measures}

% Accuracy was chosen as the sole metric for evaluating the "SOMADHAN" and "PatiGonit" datasets. This decision is grounded in the nature of math word problems, where the primary focus is on the correctness of the final answer, rather than the specific intermediate steps taken to reach that answer. Given this context, additional metrics such as precision, recall, or F1-score were deemed unnecessary for the evaluation.

For the SOMADHAN dataset, the evaluation process involved comparing the model's final answer against the correct answer. Only the accuracy of the final solution was considered relevant for assessment. On the other hand, for the PatiGonit dataset, the evaluation was based on matching the predicted equations with the actual equations, which was done manually. In both cases, the evaluation was performed through manual checks to ensure the correctness of the results, guaranteeing the reliability of the accuracy measurements for both datasets.

\subsection{Proposed Approaches}
A methodology was explored for solving Bengali math word problems, leveraging large language models (LLMs) to assess their performance and suitability for this complex task.

\subsubsection{Large-Language Models}
\begin{figure}[h]
\centering
  \includegraphics[width=1.0\columnwidth]{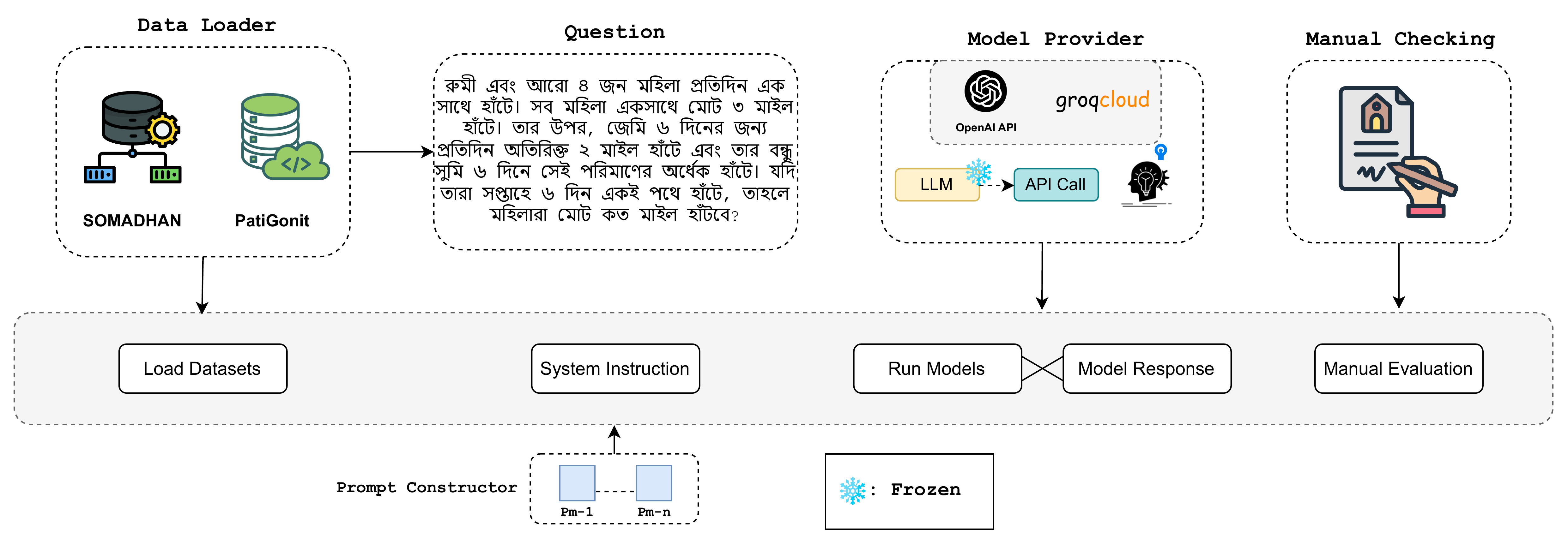}
  \caption{Schematic diagram of our proposed prompting approach}
  \label{fig:methC}
\end{figure}

When using large language models (LLMs), we evaluated their performance through in-context zero-shot and few-shot prompting, emphasizing the Chain of Thought (CoT) approach for reasoning-intensive tasks. This method involved designing prompts that guided the models to provide step-by-step explanations, enhancing clarity and interpretability. Figure~\ref{fig:methC} illustrates the schematic diagram of our proposed prompting approach.

Additionally, we applied fine-tuning to structure the CoT prompting more effectively. To balance creativity and coherence, we set the temperature to one for all models, enabling them to generate mathematical equations with detailed, logical explanations. This approach fully leveraged the models' potential while ensuring the clarity and precision necessary to solve complex mathematical problems.

\textbf{Gpt-4o and Gpt-3.5 Turbo: }GPT models, developed by OpenAI, are state-of-the-art large language models designed for a wide range of natural language processing tasks. Notable examples include GPT-4o \cite{gpt4o} with trillions of parameters and GPT-3.5 Turbo \cite{gpt3.5} with approximately 175 billion parameters. These models are known for their unparalleled ability to understand and generate human-like text, handle complex instructions, and solve intricate tasks with high precision.

Their success stems from extensive training on diverse datasets and leveraging advanced attention mechanisms to capture contextual nuances. In mathematical problem-solving, GPT models excel by employing the Chain of Thought (CoT) approach, generating detailed intermediate reasoning steps that enhance logical accuracy and interpretability. By breaking problems into smaller, manageable parts, CoT ensures systematic reasoning and correct final results. This capability makes GPT models exceptionally effective for tasks requiring deep comprehension, step-by-step execution, and transparent problem-solving processes.

\textbf{Llama (Large Language Model Meta AI): }LLaMA (Large Language Model Meta AI) is a series of advanced language models developed by MetaAI. These models\cite{Llama3}, including LLaMA-3 8B, LLaMA-3 70B, LLaMA-3.1 70B, LLaMA-3.1 8B, LLaMA-3.2 1B-preview, LLaMA-3.2 3B-preview, LLaMA-3.2 90B-text-preview, and LLaMA-3.3 70B, represent significant strides in NLP with their efficient scaling and fine-tuning capabilities. Unlike other large models, LLaMA focuses on optimizing performance through parameter-efficient designs, making them particularly effective for specific tasks even with smaller sizes. The LLaMA-3 series introduces improvements in token processing, context understanding, and task adaptability, demonstrating remarkable performance across a variety of benchmarks.

Despite having fewer parameters than GPT-4, LLaMA models excel in tasks like mathematical reasoning by leveraging optimized architectures and structured pre-training to produce accurate, coherent outputs. They effectively apply a structured reasoning approach inspired by the Chain of Thought (CoT) process, enabling step-by-step problem-solving by breaking complex tasks into intermediate steps. This systematic methodology allows LLaMA models to process and solve tasks efficiently while maintaining clarity and precision. Their ability to utilize CoT-style reasoning ensures reliable solutions and highlights their effectiveness in tackling challenges that demand logical progression and detailed analysis.

\textbf{Deepseek: }In addition to the LLaMA series, we also incorporated models from Deepseek and Qwen to further assess their performance in solving complex Bengali Math Word Problems (MWPs). The Deepseek-r1-distill-qwen-32B model \citep{Deepseek} is a distilled version of the Qwen-32B model, specifically optimized for efficient performance while reducing computational overhead. This model leverages the distillation process, which involves training a smaller model to mimic the behavior of a larger model. Through this process, the Deepseek-r1-distill-qwen-32B retains the essential capabilities of the Qwen-32B model, allowing it to perform a wide range of natural language processing (NLP) tasks, including complex problem-solving. The distillation technique enhances the model's efficiency by significantly reducing the computational resources required for inference, making it a suitable choice for tasks involving complex mathematical reasoning.

Similarly, the Deepseek-r1-distill-llama-70B model \citep{Deepseek} is a distilled variant of the LLaMA-70B model. This model is designed to achieve parameter efficiency while maintaining high levels of performance, particularly in tasks that require multi-step reasoning, such as solving math word problems. By distilling the larger LLaMA-70B model, the Deepseek-r1-distill-llama-70B maintains the essential performance characteristics of its predecessor while optimizing its computational resource usage. This balance between model performance and resource efficiency makes the Deepseek-r1-distill-llama-70B well-suited for handling the complex and computationally intensive nature of solving Bengali Math Word Problems.

\textbf{Qwen: }Additionally, we integrated Qwen models, specifically Qwen-2.5-32B \citep{Qwen} and Qwen-qwq-32B \citep{Qwen}, both of which are robust models designed to handle multi-tasking and complex reasoning. The Qwen-2.5-32B model, with its 32 billion parameters, excels in addressing a wide range of natural language processing (NLP) challenges, including multi-step reasoning and complex mathematical computations. It has demonstrated impressive generalization capabilities across various benchmarks, particularly in tasks that require structured problem-solving. This model's architecture is specifically optimized to manage intricate problem-solving tasks, making it well-suited for Bengali Math Word Problems (MWPs).

Meanwhile, the Qwen-qwq-32B model further enhances the performance of the Qwen-2.5 by incorporating advanced optimizations that enhance its efficiency in solving mathematical problems. These optimizations improve the model’s ability to handle tasks requiring deep reasoning and multi-step calculations, making it more efficient and accurate in complex problem-solving scenarios. Together, these models represent the latest advancements in large language models, offering significant improvements in the Bengali MWP-solving framework and contributing meaningfully to its effectiveness in solving computationally challenging problems.

\subsection{Prompting Techniques}
While working with large language models (LLMs), we aimed to make our instructions clear and easy to follow so the models could better understand our tasks and give accurate responses. This was important because LLMs often generate different answers depending on how the prompts are designed. Creating these prompts was a step-by-step process with challenges due to the way information is represented differently across models.
Our study focused on improving instructions and prompts to guide the models in acting like math instructors and solving math problems with accurate solutions. We also created three sets of instructions and developed unique prompt styles for our proposed dataset. Effective prompting is particularly important for encouraging the models to adopt a Chain of Thought (CoT) approach. Chain-of-thought (CoT) prompting has several key advantages for improving how language models reason:

\begin{enumerate}
    \item Helps models break complex problems into smaller, easier steps, allowing them to focus on tasks that need more reasoning.

    \item Gives us a clearer view of how the model arrives at an answer, making it easier to spot where the reasoning went wrong (though fully explaining a model’s thinking is still challenging).

    \item Can be used for various tasks, like solving math problems, understanding everyday situations, and manipulating symbols, and can potentially apply to any task that humans can solve using language.

    \item CoT reasoning can be triggered in large language models by simply including examples of reasoning steps in a few-shot prompt.
    
\end{enumerate}

\subsubsection{Zero Shot Prompting}
\begin{figure}[!htbp]
\centering
\begin{minipage}{1.0\textwidth}
  \centering
\includegraphics[width=0.9\textwidth]{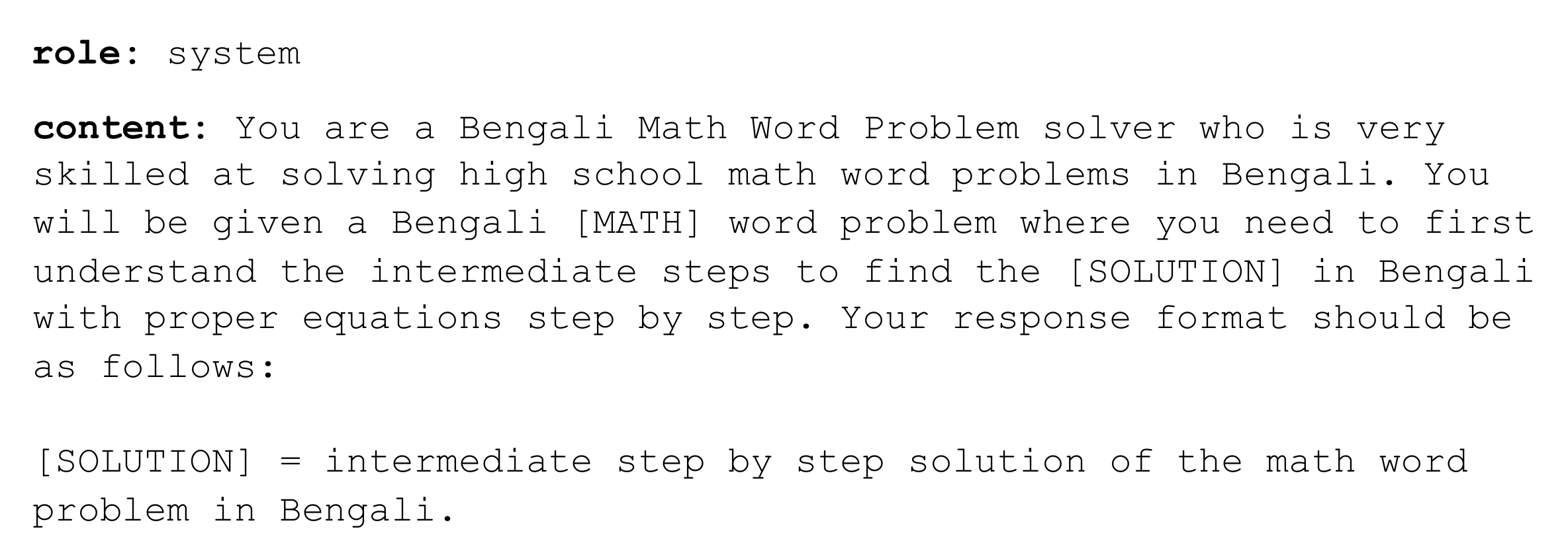}
\subcaption{System Instruction}\label{fig:prompt1}
\end{minipage}%

\begin{minipage}{1.0\textwidth}
  \centering
\includegraphics[width=0.92\textwidth]{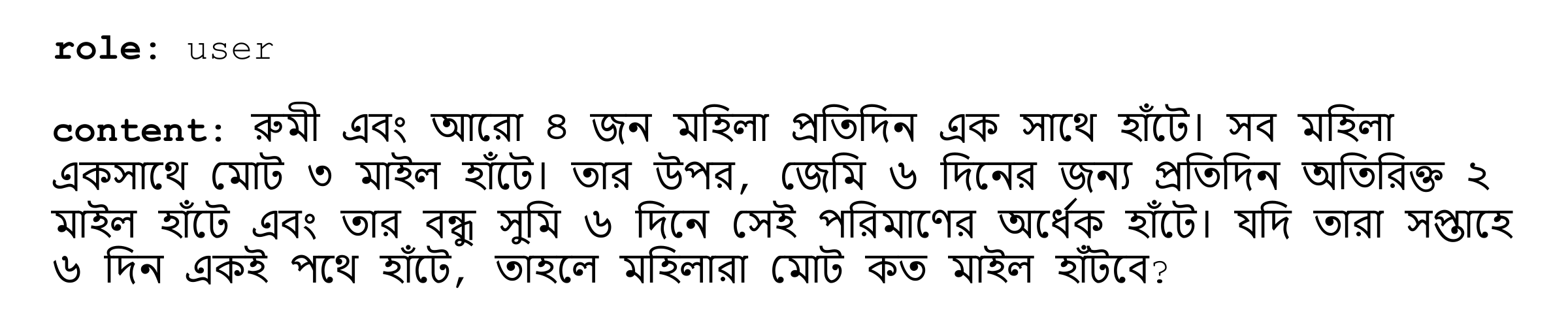}
\subcaption{User Question}\label{fig:prompt2}
\end{minipage}%

\caption{Instruction (Prompt-1) for SOMADHAN Dataset (Zero Shot Prompting)} \label{fig:promptzero}
\end{figure}

\begin{figure}[!htbp]
\centering
\begin{minipage}{1.0\textwidth}
  \centering
\includegraphics[width=0.9\textwidth]{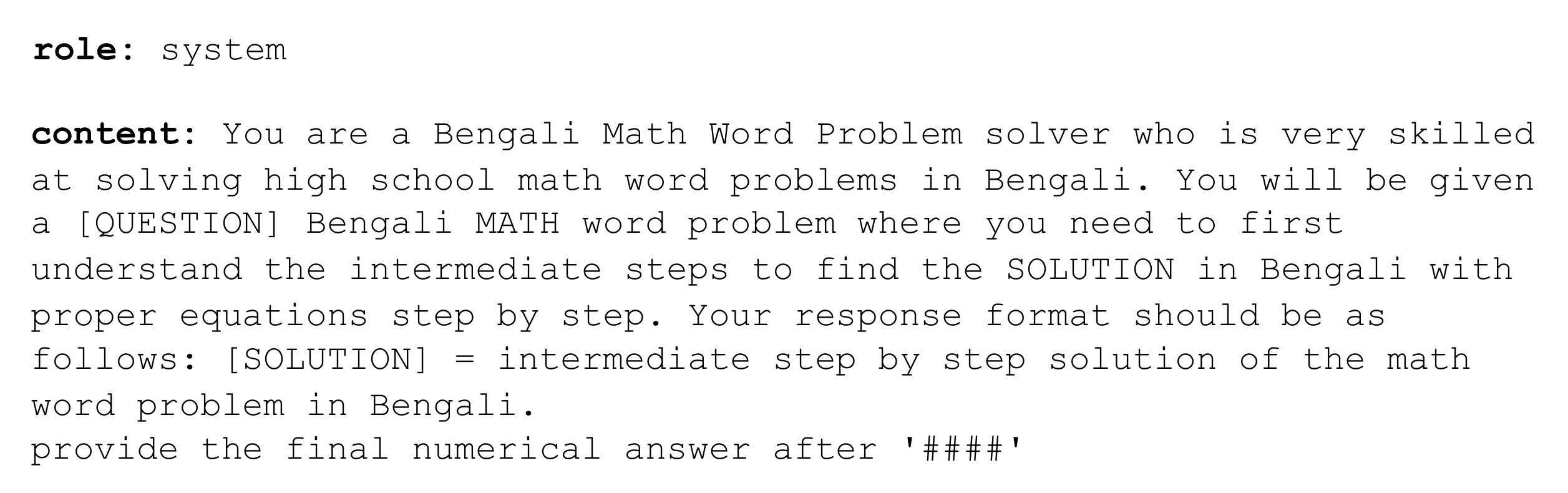}
\subcaption{System Instruction}\label{fig:prompt1}
\end{minipage}%

\begin{minipage}{1.0\textwidth}
  \centering
\includegraphics[width=0.92\textwidth]{Prompt/zerouser.pdf}
\subcaption{User Question}\label{fig:prompt2}
\end{minipage}%

\caption{Instruction (Prompt-2) for SOMADHAN Dataset (Zero Shot Prompting)} \label{fig:promptzero2}
\end{figure}

Zero-shot prompting is particularly useful for testing a model's inherent capabilities and adaptability without the need for extra fine-tuning or example-based context. In our approach, we applied zero-shot prompting for both Chain-of-Thought (CoT) and standard prompting. For CoT, we provided clear chain-of-thought instructions that described the task and outlined the expected output, enabling the models to generate intermediate reasoning steps autonomously. This method encourages the LLMs to create their own context and refine their reasoning process, ultimately leading to more accurate and logical results. In contrast, for standard prompting, no prior instructions on solving the problem were given. The model was simply presented with the question, and it was expected to generate the answer without any guidance on breaking down the problem. To enhance the reasoning process, two separate system instructions were incorporated in our prompting techniques. Figure~\ref{fig:promptzero} and Figure~\ref{fig:promptzero2} shows the two types of system instruction (Prompt-1 \& Prompt-2) for CoT with zero shot Prompting. Moreover, Figure~\ref{fig:stdprompt} illustrates the system instruction that we have used for the standard prompting.

\subsubsection{Few Shot Prompting}
\begin{figure}[!htbp]
\centering
\begin{minipage}{0.9\textwidth}
  \centering
\includegraphics[width=1.0\textwidth]{Prompt/zeroshot.pdf}
\subcaption{System Instruction}\label{fig:prompt01}
\end{minipage}%

\begin{minipage}{0.9\textwidth}
  \centering
\includegraphics[width=1.0\textwidth]{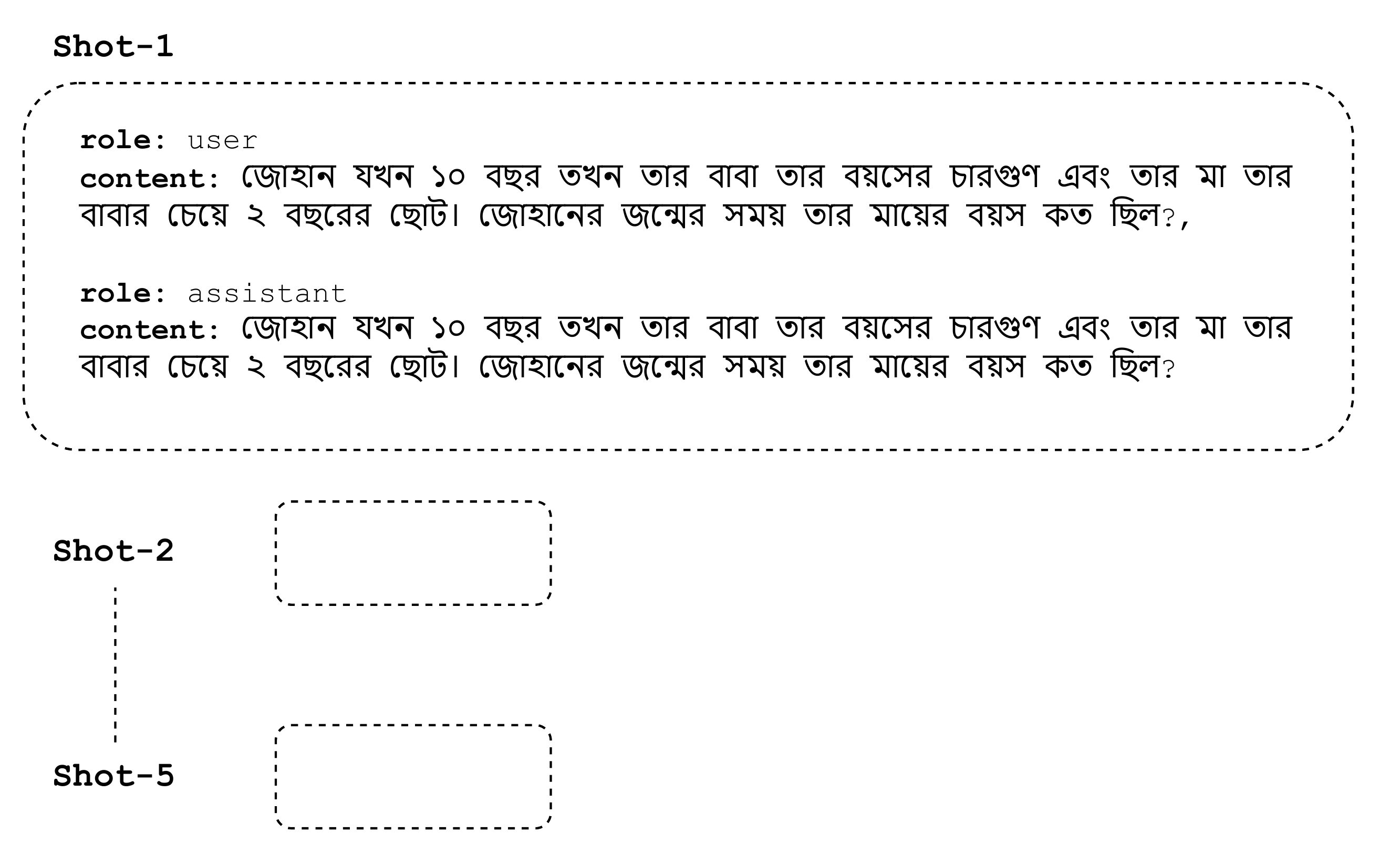}
\subcaption{5 Shot Prompting}\label{fig:prompt02}
\end{minipage}%

\begin{minipage}{0.9\textwidth}
  \centering
\includegraphics[width=1.0\textwidth]{Prompt/zerouser.pdf}
\subcaption{User Question}\label{fig:prompt03}
\end{minipage}%

\caption{Instruction (Prompt-1) for SOMADHAN Dataset (Few (5) Shot Prompting)} \label{fig:promptfew1}
\end{figure}

\begin{figure}[!htbp]
\centering
\begin{minipage}{0.9\textwidth}
  \centering
\includegraphics[width=1.0\textwidth]{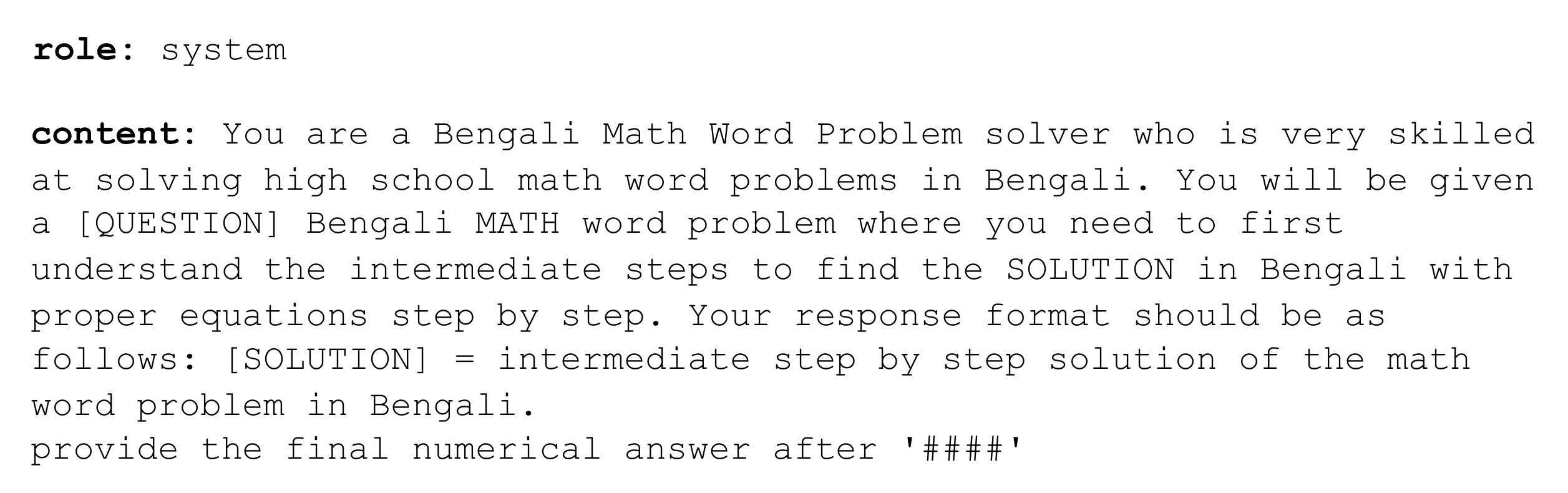}
\subcaption{System Instruction}\label{fig:prompt01}
\end{minipage}%

\begin{minipage}{0.9\textwidth}
  \centering
\includegraphics[width=1.0\textwidth]{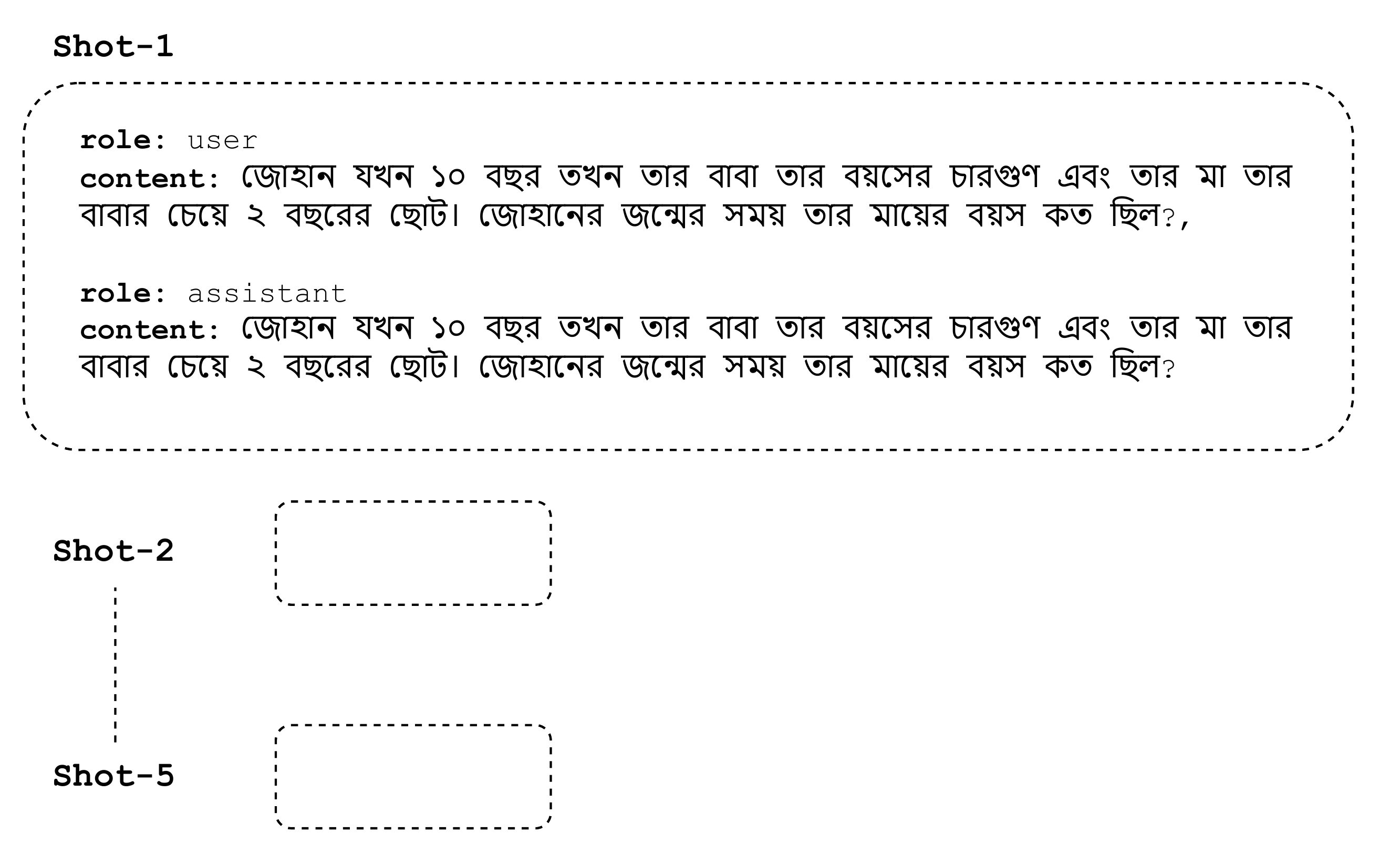}
\subcaption{5 Shot Prompting}\label{fig:prompt02}
\end{minipage}%

\begin{minipage}{0.9\textwidth}
  \centering
\includegraphics[width=1.0\textwidth]{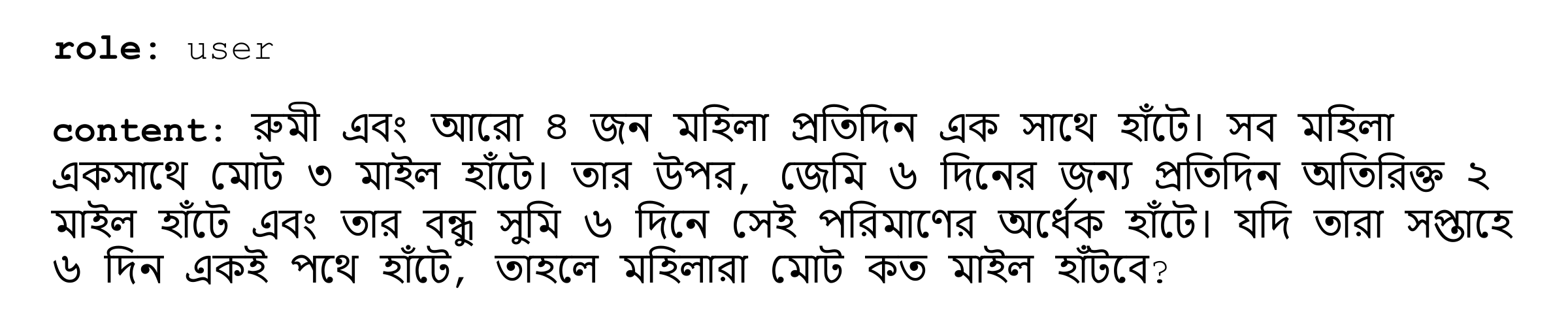}
\subcaption{User Question}\label{fig:prompt03}
\end{minipage}%

\caption{Instruction (Prompt-2) for SOMADHAN Dataset (Few (5) Shot Prompting)} \label{fig:promptfew2}
\end{figure}

\begin{figure}[!htbp]
\centering
\begin{minipage}{1.0\textwidth}
  \centering
\includegraphics[width=0.9\textwidth]{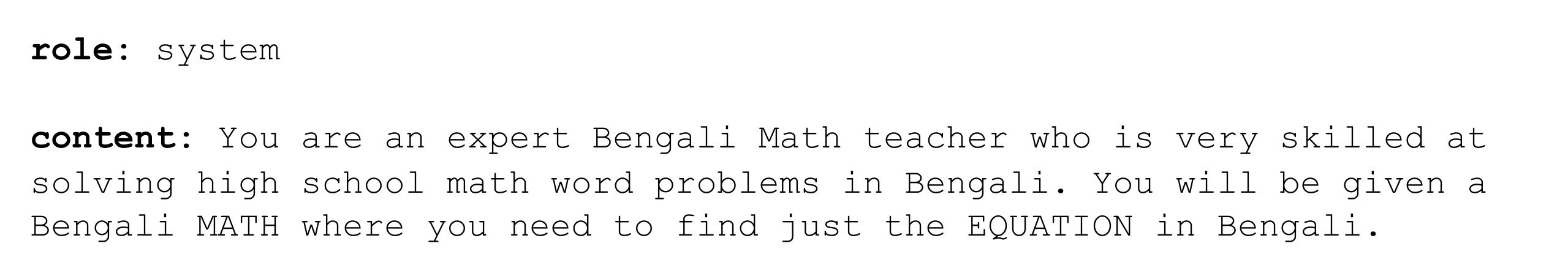}
\subcaption{System Instruction}\label{fig:prompt01}
\end{minipage}%

\begin{minipage}{1.0\textwidth}
  \centering
\includegraphics[width=0.9\textwidth]{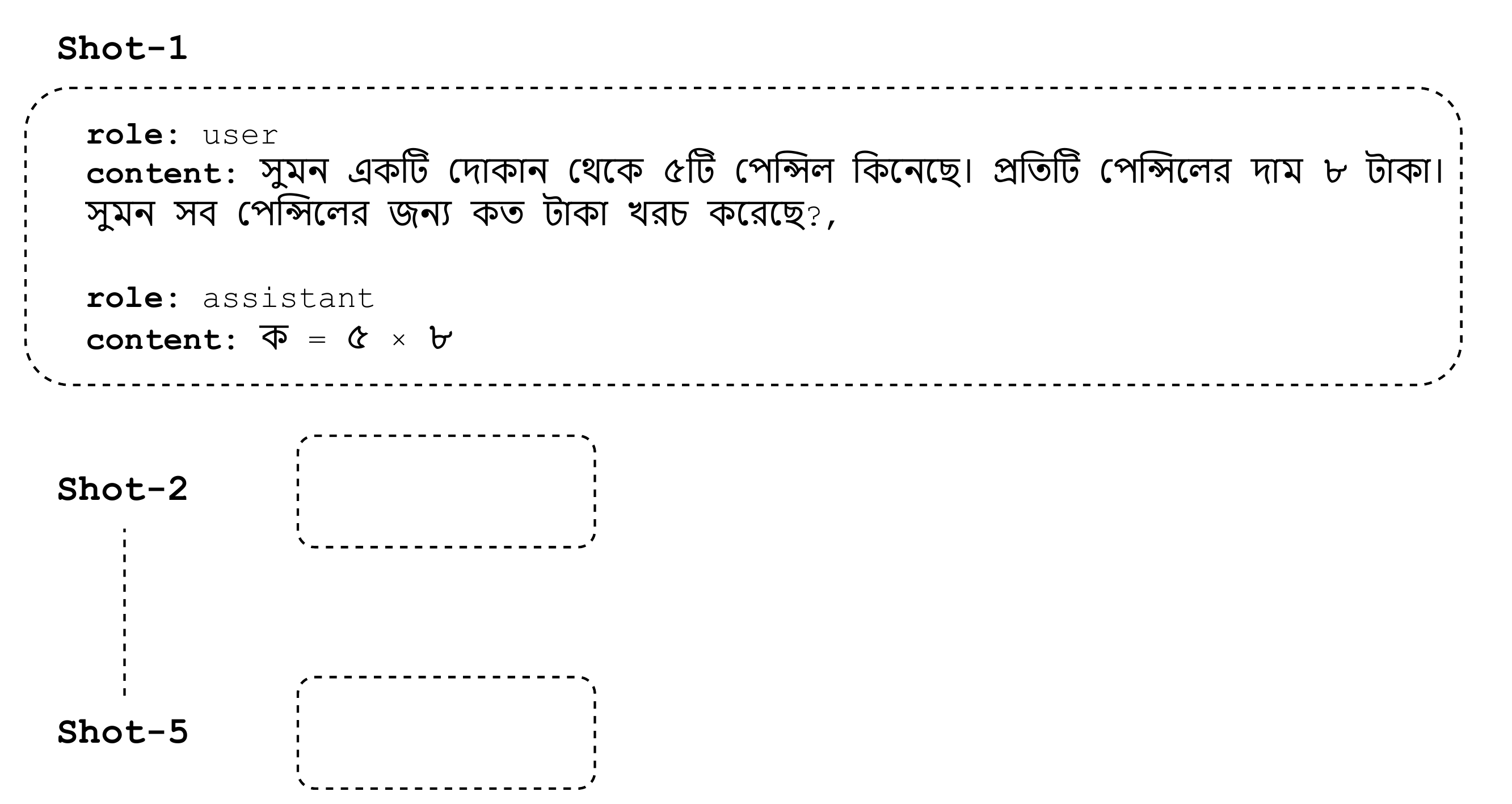}
\subcaption{5 Shot Prompting}\label{fig:prompt02}
\end{minipage}%

\begin{minipage}{1.0\textwidth}
  \centering
\includegraphics[width=0.9\textwidth]{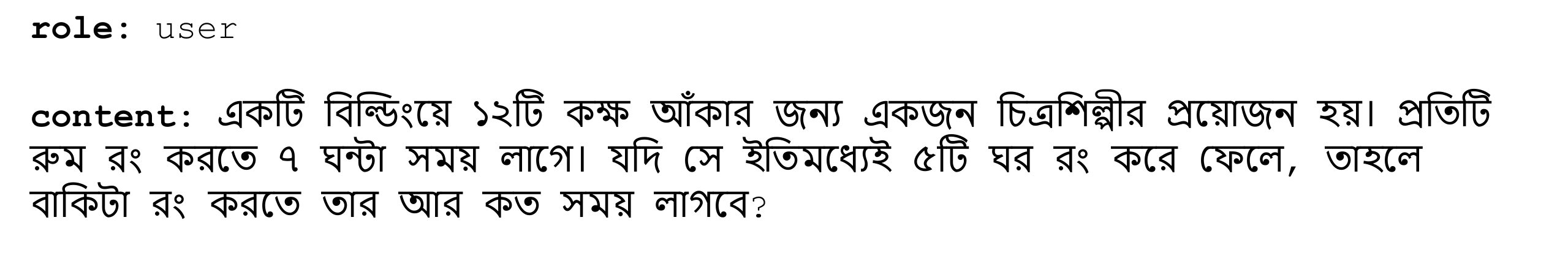}
\subcaption{User Question}\label{fig:prompt03}
\end{minipage}%

\caption{Instruction (Prompt-3) for PatiGonit Dataset (Few (5) Shot) Prompting} \label{fig:promptfew3}
\end{figure}

\begin{figure}[!htbp]
  \includegraphics[width=1.0\columnwidth]{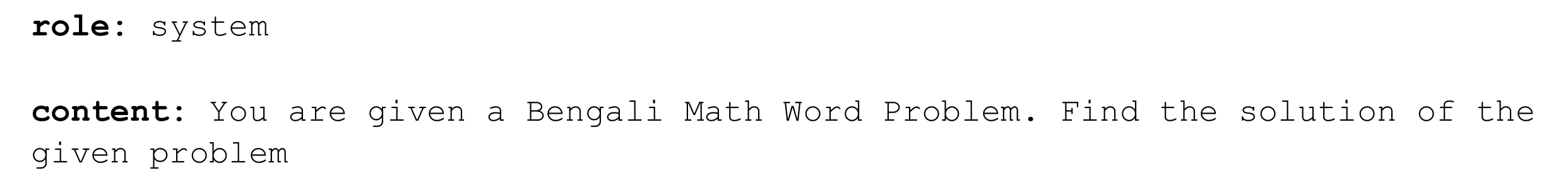}
  \caption{System Instruction for standard prompting}
  \label{fig:stdprompt}
\end{figure}

In our approach, we focused on leveraging few-shot learning, which has been shown to outperform zero-shot learning, as demonstrated in studies by \cite{2} and \cite{5}. Unlike zero-shot, where no prior examples are provided, few-shot learning uses a small number of example prompts to guide the model’s reasoning process. We conducted experiments with GPT-4, GPT-3.5, the LLaMA-3 series, Deepseek, and Qwen models, utilizing training data to enhance their performance. To optimize API costs, we manually included five few-shot prompts. For our two datasets, we designed distinct prompts. Several key steps were taken: we taught the model how to approach and reason through the task, followed by question-and-answer prompts for generating answers. In contrast to zero-shot prompting, which solely relies on the model’s pre-existing knowledge without examples, our few-shot approach provides explicit examples of the task, allowing the model to adapt to the specific problem structure. Finally, math problem tests were conducted to assess the model’s performance. Our method further augments each few-shot example by incorporating a chain of thought, improving accuracy and clarity in the results. In Figure~\ref{fig:promptfew1},~\ref{fig:promptfew2}, and ~\ref{fig:promptfew3} we showcase examples of few-shot prompts for both of the datasets. Additionally, Figure~\ref{fig:stdprompt} illustrates the system instruction that we have used for the standard prompting.

\subsection{Fine Tuning}
To further improve model performance on complex Bengali math word problems, we explore fine-tuning techniques tailored to reasoning-based tasks. In this section, we describe two approaches—standard fine-tuning using OpenAI’s GPT-3.5 and parameter-efficient fine-tuning using Low-Rank Adaptation (LoRA)—both enhanced with chain-of-thought supervision.

\subsubsection{Gpt-3.5 Standard Tuning}
Fine-tuning \cite{finetune} enhances few-shot learning by training on a larger set of examples, resulting in improved performance across various tasks. In this approach, we integrated chain of thought to encourage models to generate step-by-step reasoning in their solutions. Once a model is fine-tuned, fewer examples are required in the prompt, which reduces costs and enables faster response times. The fine-tuning process consists of the following steps, which differ from few-shot prompting:

\begin{enumerate}
    \item Compile and upload the required data for training, ensuring the inclusion of chain-of-thought examples where necessary
    \item Develop a new model that is optimized for better performance, focusing on improving reasoning abilities with intermediate steps
    \item Analyze the outcomes and make necessary adjustments to ensure that the model produces clear, logical reasoning along with its final answers
    \item Utilize the refined and optimized model, leveraging the chain of thought to maintain accurate, interpretable problem-solving
\end{enumerate}

\begin{figure}[!htbp]
  \centering
  \setlength{\fboxrule}{0.5mm} % Line thickness
  \setlength{\fboxsep}{2mm} % Padding between image and box
  \fbox{\includegraphics[width=0.9\columnwidth]{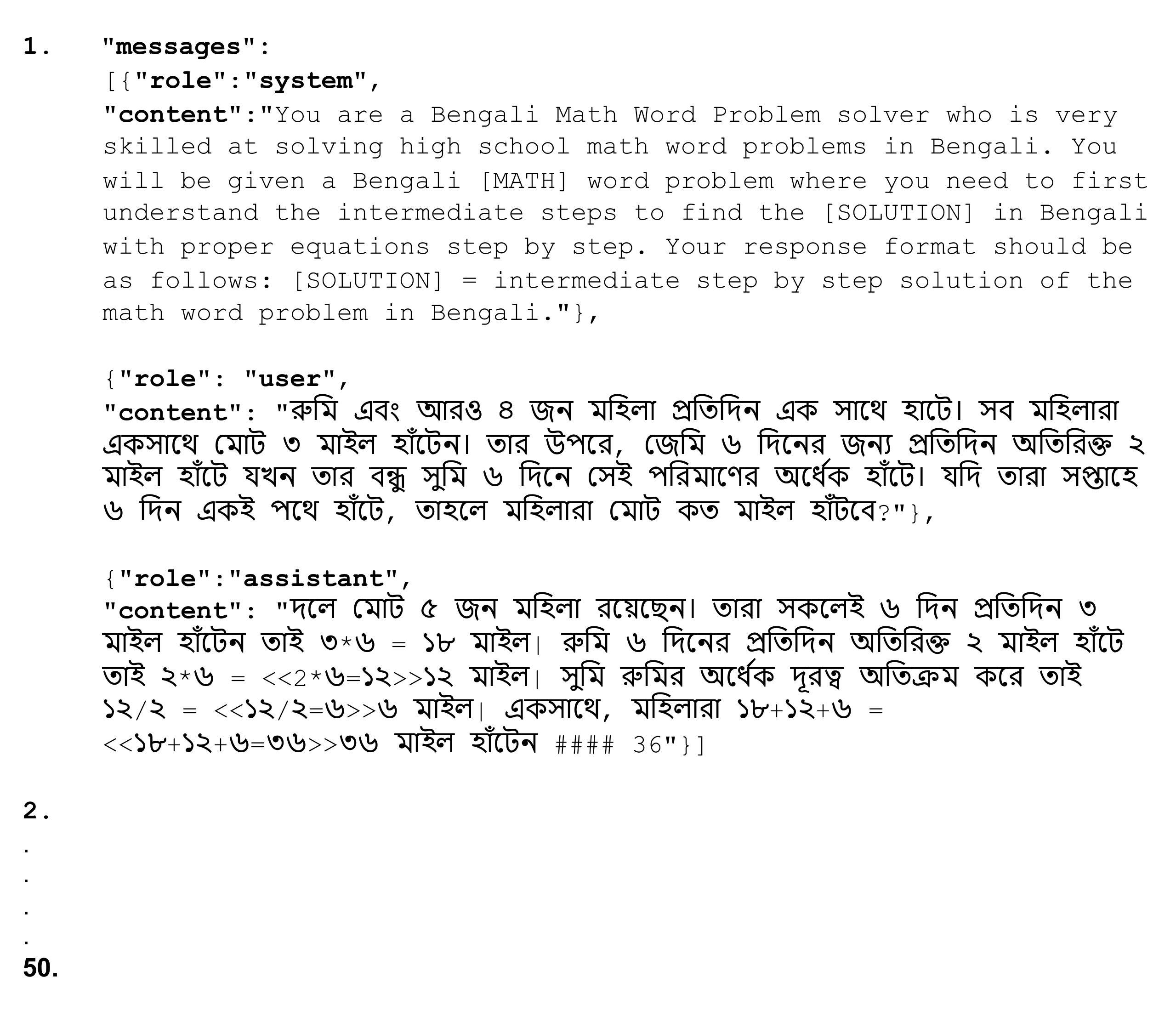}} % Image inside dotted box
  \caption{Structure of JSON file format}
  \label{fig:json}
\end{figure}

We chose to fine-tune the recommended model, GPT-3.5-turbo-0125. During this process, we varied only the number of epochs to optimize the model’s performance.
To build the fine-tuned model, we used 50 examples with appropriate instructions, including explicit chain-of-thought reasoning for complex problems. This ensured the model not only provided answers but also demonstrated the thought process behind solving math problems. The examples followed the JSONL file format as outlined in the OpenAI API documentation. Figure~\ref{fig:json} shows an example of how the JSON file was structured to create the fine-tuned model.

Our SOMADHAN dataset required proper chain-of-thought reasoning to produce correct solutions, so fine-tuning was essential. The models were trained to break down complex problems into intermediate steps, ensuring that the reasoning behind the answers was explicit and interpretable. In contrast, the PatiGonit dataset only contains math word problem equations, which require less complex reasoning. For this dataset, GPT-3.5 Turbo and GPT-4 models performed well with few-shot prompting, so extensive fine-tuning was not necessary.

\subsubsection{LoRA Fine Tuning}

\begin{figure}[!htbp]
\centering
  \includegraphics[width=0.5\columnwidth]{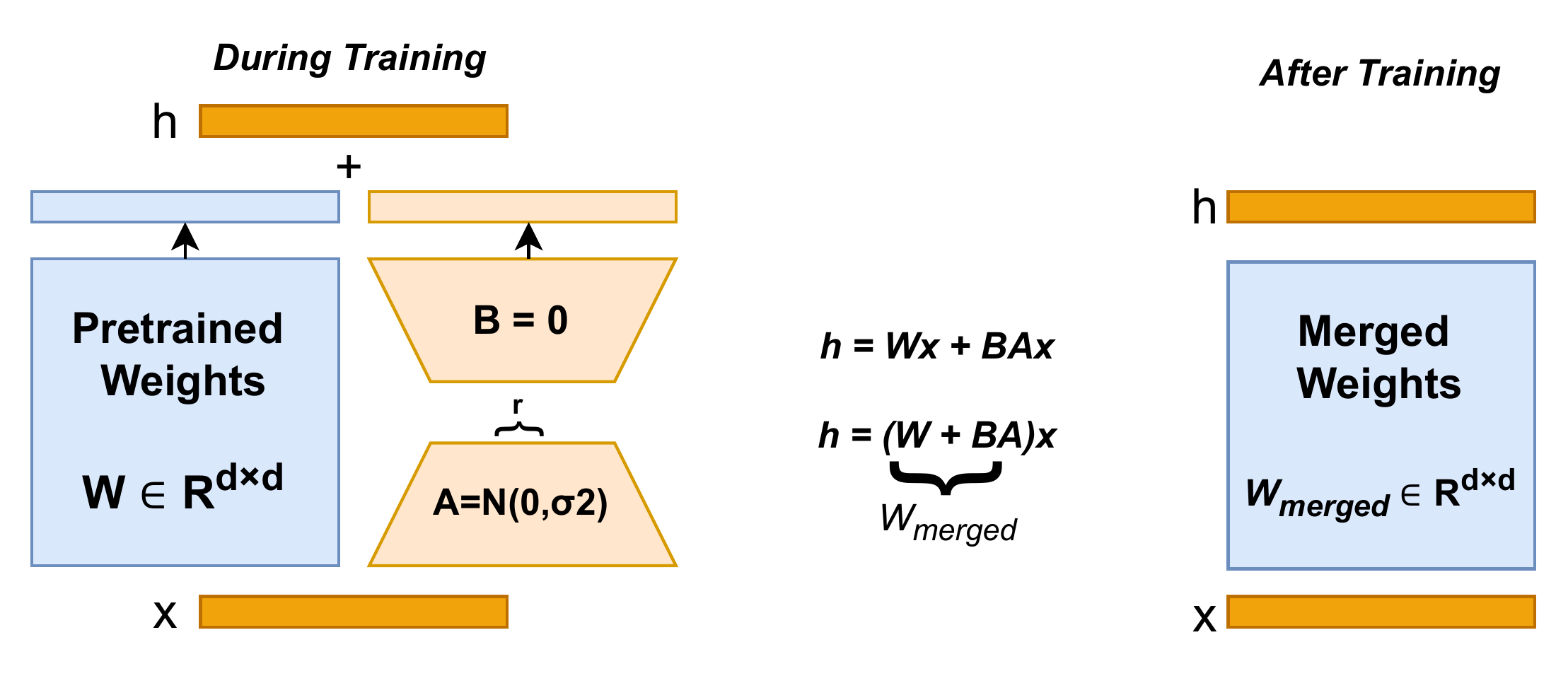}
  \caption{LoRA Fine-Tuning Process: During and After Training (LoRA, 2024)\cite{LoRA}}
  \label{fig:lora}
\end{figure}

For SOMADHAN dataset, we employed Low-Rank Adaptation (LoRA) \cite{1} for fine-tuning. LoRA improves fine-tuning by learning two smaller weight-updating matrices, allowing the large model to remain frozen while only the smaller matrices are updated with new data. After training, the updated matrices are recombined with the original weights to create the final model. This method significantly reduces the number of parameters and accelerates the fine-tuning process while saving storage. As part of this process, we also integrated chain-of-thought examples into the fine-tuning dataset to encourage the model to generate logical, step-by-step reasoning. Figure~\ref{fig:lora}
shows an overview of the process.

\section{Experimental Setup}

To evaluate the performance of large language models on Bengali Math Word Problems, we designed a comprehensive experimental setup. This includes configuring the computational environment, organizing train-test splits for both datasets, and defining evaluation metrics tailored to reasoning-based and equation-based tasks.

Our work was conducted on Google Colab Notebook with Python 3.10.12, PyTorch 2.0.1, a Tesla T4 GPU (15 GB), 12.5 GB of RAM, and 64 GB of disk space.

\subsection{Train-Test-Validation splits}
For our SOMADHAN dataset’s prompting and evaluation, we divided the dataset into two parts: 80\% for testing and 20\% for training. Specifically, 3200 examples were assigned for testing, while 800 examples were reserved for training. Prompting and fine-tuning samples were taken from the training set, and evaluation was done using the testing set. The same splitting technique was applied for the PatiGonit dataset. However, due to token costs, only 1000 samples from the testing set were considered for evaluation in our experiments for both datasets. This decision allowed us to manage resource constraints effectively while still obtaining valuable insights into model performance.

\subsection{Evaluation metrics}
\textbf{SOMADHAN Dataset Evaluation}

The model's output can be represented as:
\begin{center}
    f(x) = ($s_{1}$, $s_{2}$, ...... ,$s_{k}$, $r_{pred}$)
\end{center}
where $s_{1}$, $s_{2}$, ...... ,$s_{k}$ are the intermediate steps and $r_{pred}$ is the predicted final result.

For evaluation, the intermediate reasoning steps $s_{1}$, $s_{2}$, ...... ,$s_{k}$ are not considered. Only the final result $r_{pred}$ is compared with the ground truth $r_{true}$:

\begin{equation}
    \text{Evaluation}(f(x), r_{\text{true}}) = 
    \begin{cases}
        1 & \text{if } r_{\text{pred}} = r_{\text{true}}, \\
        0 & \text{if } r_{\text{pred}} \neq r_{\text{true}}.
    \end{cases}
\end{equation}

The model is considered to have successfully solved the problem if: $r_{\text{pred}} = r_{\text{true}}$

If the dataset contains \textit{N} problems {$x_{1}$, $x_{2}$, ...... , $x_{\textit{N}}$}, the accuracy \textit{A} is computed as:
\begin{center}
   A = $\frac{1}{N}$ $\sum_{i=1}^{N}$ $\mathbf{1}(r_{\text{pred}, i} = r_{\text{true}, i})$ 
\end{center}

where $\mathbf{1}(\cdot)$ is the indicator function, which equals 1 if the condition is true and 0 otherwise.

\textbf{PatiGonit Dataset Evaluation}

We used equation accuracy, which compares the predicted equation with the correct equation rather than focusing on the final solution. This metric more effectively measures the model's ability to generate the correct equation, regardless of minor variations in the numerical result. By evaluating the structure of the equation itself, this method ensures that the model is solving the problem with the right logical steps, even if the final solution is not exactly the same.

The model's output can be represented as:
\begin{center}
f(x) = ($e_{\text{pred}}$)
\end{center}
where $e_{\text{pred}}$ is the equation generated by the model.

Evaluation focuses on comparing the predicted equation $e_{\text{pred}}$ with the ground truth equation $e_{\text{true}}$. This approach ensures that the model generates the correct mathematical equation, which is the primary focus of our evaluation.

% We used solution accuracy, which compares the final solution from the predicted equation with the correct answer. This metric more effectively measures the model's ability to solve math word problems by focusing on the correctness of the numerical result, regardless of minor variations in equation structure.

% Chapter~\ref{chap:intro} problem statement states that, The model's output can be represented as:
% \begin{center}
%     f(x) = (e, $r_{\text{pred}}$)
% \end{center}
% where $e$ is the equation generated by the model, and $r_{\text{pred}}$ is the predicted final result.

% Evaluation focuses on comparing $r_{\text{pred}}$ with the ground truth $r_{\text{true}}$, while optionally verifying the correctness of the generated equation $e$.

\section{Result Analysis}

The result analysis section provides a detailed evaluation of the model's performance on the SOMADHAN and PatiGonit datasets. We examined the accuracy to evaluate solution correctness to assess the effectiveness of our proposed approaches. 

\subsection{Chain of Thoughts based results}

The results in Table~\ref{tab:my-table1} and !\ref{tab:my-table2} show key distinctions between Standard prompting and Chain of Thought (CoT) prompting, highlighting their respective strengths and weaknesses. In Table~\ref{tab:my-table1}, the performance of six different large language models (LLMs) is compared across the SOMADHAN and PatiGonit datasets using both Standard prompting and Chain of Thought (CoT) prompting, with evaluations based on Zero-Shot and Few-Shot settings. Prompt Style-1 was used for the experiments in this table. The models include GPT-3.5, GPT-4o, and various versions of the Llama series.

% \usepackage{multirow}
\begin{table}[h]
\centering
\renewcommand{\arraystretch}{1.5}
\caption{Performance comparison of Chain of Thoughts (CoT) prompting (Prompt-1) versus Standard prompting for various large language models on the "SOMADHAN" dataset. A benchmark has been set for the "PatiGonit" dataset using Few Shot prompting (Prompt-3) technique. All metrics are Accuracy (\%). (*) represents that the accuracy is correct but the responses of the model was in English.}
\label{tab:my-table1}
\footnotesize
\begin{tabular}{l l cccc c }
\hline
\multirow{3}{*}{\textbf{Models}}                             & \multirow{3}{*}{\textbf{Parameters}} & \multicolumn{4}{c}{\textbf{SOMADHAN}}                                                                                                                   & \textbf{PatiGonit}                                                                \\ \cline{3-6} 
                                                             &                                      & \multicolumn{2}{c}{\textbf{Standard}}                                               & \multicolumn{2}{c}{\textbf{Chain of Thoughts (CoT)}}                               & \multirow{2}{*}{\textbf{\begin{tabular}[c]{@{}c@{}}------------------\\Few (5) Shot\end{tabular}}} \\ \cline{3-6}
                                                             &                                      & \multicolumn{1}{c}{\textbf{Zero Shot}} & \multicolumn{1}{c}{\textbf{Few (5) Shot}} & \multicolumn{1}{c}{\textbf{Zero Shot}} & \textbf{Few (5) Shot} &                                                                                   \\ \hline
\begin{tabular}[c]{@{}c@{}}GPT-3.5\end{tabular} & 175B                                 & \multicolumn{1}{c}{23.0}                  & \multicolumn{1}{c}{24.0}                     & \multicolumn{1}{c}{$24.0_{ (+1.0)}$}                  & $\textbf{30.0}_{(+6.0)}$                    & 86.0                                                                               \\ 
GPT-4o                                                       & 1.3T(approx.)                                    & \multicolumn{1}{c}{79.0}                  & \multicolumn{1}{c}{80.0}                     & \multicolumn{1}{c}{$80.0_{(+1.0)}$}                  & $\textbf{83.0}_{(+3.0)}$                     & \textbf{99.0}                                                                                 \\ \hline
\multirow{2}{*}{Llama-3}                                     & 70B                                  & \multicolumn{1}{c}{\textbf{70.0}*}                  & \multicolumn{1}{c}{67.0}                     & \multicolumn{1}{c}{$62.0_{(-8.0)}$}                  & $65.0_{(-2.0)}$                     & 96.0                                                                                 \\  
                                                             & 8B                                   & \multicolumn{1}{c}{\textbf{47.0}*}                  & \multicolumn{1}{c}{33.0}                     & \multicolumn{1}{c}{$24.0_{(-23.0)}$}                  & $19.0_{(-14.0)}$                     & 94.0                                                                                 \\ \hline
Llama-3.1                                                    & 70B                                  & \multicolumn{1}{c}{78.0}                  & \multicolumn{1}{c}{70.0}                     & \multicolumn{1}{c}{$\textbf{80.0}_{(+2.0)}$}                  & $70.0_{(+0.0)}$                     & 96.0                                                                                \\ 
Llama-3.2                                                    & 90B                                  & \multicolumn{1}{c}{75.0}                  & \multicolumn{1}{c}{72.0}                     & \multicolumn{1}{c}{$\textbf{79.0}_{(+4.0)}$}                  & $69.0_{(-3.0)}$                     & 96.0     \\
Llama-3.3    &70B        &\multicolumn{1}{c}{76.0}  & \multicolumn{1}{c}{78.0}                     & \multicolumn{1}{c}{$80.0_{(+4.0)}$}                  & $\textbf{87.0}_{(+9.0)}$                     & 96.0  \\ \hline
\end{tabular}
\end{table}

The GPT-3.5 model achieves notable results in both zero-shot and few-shot settings. On the SOMADHAN dataset, it reaches 23.0\% accuracy in zero-shot and 24.0\% accuracy in few-shot using standard prompting. However, with Chain of Thought (CoT) prompting, GPT-3.5's performance improves, reaching 24.0\% accuracy in zero-shot and \textbf{30.0\%} in few-shot, which is a significant increase of 6.0\%. This indicates the effectiveness of the CoT approach for handling more complex tasks. It also performs decently in the PatiGonit dataset, achieving 86.0\% with few-shot prompting. GPT-4o, with approximately 1.3 trillion parameters, demonstrates substantial improvements over GPT-3.5. For SOMADHAN, the model reaches 79.0\% in zero-shot and 80.0\% in few-shot using standard prompting. With CoT prompting, its performance dramatically improves, achieving 80.4\% in zero-shot and \textbf{83.0\%} in few-shot, marking an impressive 3.0\% increase in the few-shot setting. This shows that GPT-4o can handle complex reasoning tasks effectively with CoT. In the PatiGonit dataset, GPT-4o achieves \textbf{99.0\%} accuracy in few-shot, nearly perfect, emphasizing its advanced capabilities.

In the case of Llama-3 models with 70B and 8B parameters, the performance differs significantly between standard prompting and Chain of Thought (CoT) prompting. For Llama-3 (70B), the highest accuracy was achieved with standard prompting in the zero-shot setting, reaching 76\% accuracy. However, it is important to note that all responses were in English, even though the dataset required responses in Bengali. When CoT prompting was applied, the accuracy dropped slightly to 65\% in zero-shot and few-shot settings. Despite the decrease in accuracy, the responses shifted to Bengali, indicating that while CoT improved the language adaptation, it did not significantly improve the accuracy when compared to standard prompting. For Llama-3 (8B), the performance was considerably lower due to the reduced parameter size. The highest accuracy for Llama-3 (8B) was 47\% using standard prompting in zero-shot, but when CoT was applied, the accuracy further decreased to 24\% in zero-shot and 19\% in few-shot. This suggests that smaller models like Llama-3 (8B) do not benefit as much from CoT prompting in case of Bengali as larger models do. In this case, CoT did not help improve accuracy significantly, and the language shift to Bengali did not overcome the challenges posed by the small parameter size. 

For Llama-3.1 (70B), the highest accuracy was achieved using Chain of Thought (CoT) prompting with zero-shot, where it reached \textbf{80\%}. This represents a notable improvement of 2\% over the standard zero-shot prompting, showcasing the positive impact of CoT on the model’s performance. However, when few-shot prompting was applied along with CoT, the performance remained the same as with standard few-shot prompting, indicating that CoT did not contribute additional improvements in the few-shot setting for this model. For Llama-3.2 (90B), the best accuracy was also achieved with Chain of Thought prompting in the zero-shot setting, where it attained \textbf{79\%}. This marked a 4\% increase in accuracy compared to standard zero-shot prompting, highlighting the benefit of CoT for this model. However, when few-shot prompting was used in combination with CoT, the accuracy declined by 3\% compared to standard few-shot prompting, suggesting that while CoT improved performance in the zero-shot setting, it did not have a similar positive effect in the few-shot setting for this model. For Llama-3.3 (70B), the highest accuracy of \textbf{87\%} was achieved on the Somadhan dataset, outperforming the other models. Without the application of Chain of Thought (CoT), the model achieved its highest accuracy of 78\% using standard prompting. However, when CoT was applied with few-shot prompting, the model’s performance significantly improved by 9\%, reaching the peak accuracy of 87\%. This demonstrates the considerable enhancement that CoT along with few shot can provide, especially in tasks involving more complex reasoning, and highlights Llama-3.3 as the top-performing model in this comparison.

In particular, the Llama-3 (70B) and Llama-3 (8B) models exhibited a noticeable drop in performance when using CoT for Bengali language tasks. Despite their strong performance in English, the models struggled to achieve comparable results in Bengali, likely due to the complexity of language-specific reasoning in CoT and their relatively smaller parameter sizes. While CoT improved the performance for larger models like GPT-4o, the benefits for Llama-3 models were more modest, especially in few-shot settings. This disparity highlights the importance of model size and prompt style in achieving optimal performance. To address these challenges, we decided to apply a different prompt style (Prompt Style 2) on the same dataset to investigate whether it could provide a more effective approach, particularly for the Llama models when handling Bengali language tasks. This experiment aims to explore potential improvements and provide better insights into the effectiveness of CoT in multilingual settings.

% \usepackage{multirow}
\begin{table}[h]
\centering
\renewcommand{\arraystretch}{1.2}
\caption{Performance comparison of Chain of Thoughts (CoT) prompting (Prompt-2) versus Standard prompting for various large language models on the "SOMADHAN" dataset. All metrics are Accuracy (\%). (*) represents that the accuracy is correct but the responses of the model was in English.}
\label{tab:my-table2}
\begin{tabular}{c c cc cc}
\toprule
\multirow{2}{*}{\textbf{Models}} & \multirow{2}{*}{\textbf{Parameters}} & \multicolumn{2}{c}{\textbf{Standard}}                                               & \multicolumn{2}{c}{\textbf{Chain of Thoughts (CoT)}}                                \\ \cline{3-6} 
                                 &                                      & \multicolumn{1}{c}{\textbf{Zero Shot}} & \multicolumn{1}{c}{\textbf{Few (5) Shot}} & \multicolumn{1}{c}{\textbf{Zero Shot}} & \multicolumn{1}{c}{\textbf{Few (5) Shot}} \\ \midrule
deepseek-r1-distill-qwen         & 32B                                  & \multicolumn{1}{c}{39.0*}                 & 44.0*                                         & \multicolumn{1}{c}{$48.0^*_{(+9.0)}$}                 & $\textbf{51.0*}_{(+7.0)}$                                         \\ 
deepseek-r1-distill-llama        & 70B                                  & \multicolumn{1}{c}{56.0}                 & 60.0                                         & \multicolumn{1}{c}{$62.0_{(+6.0)}$}                 & $\textbf{66.0}_{(+6.0)}$                                         \\ \hline
qwen-2.5                         & 32B                                  & \multicolumn{1}{c}{51.0}                 & 68.0                                         & \multicolumn{1}{c}{$70.0_{(+19.0)}$}                 & $\textbf{71.0}_{(+3.0)}$                                         \\ 
qwen-qwq                         & 32B                                  & \multicolumn{1}{c}{24.0*}                 & 29.0*                                         & \multicolumn{1}{c}{$33.0*_{(+9.0)}$}                 & $\textbf{35.0*}_{(+6.0)}$                                         \\ \hline
\multirow{2}{*}{llama3}          & 8B                                   & \multicolumn{1}{c}{\textbf{47.0}*}               & 33.0                                       & \multicolumn{1}{c}{$36.0_{(-11.0)}$}                 & $33.0_{(+0.0)}$                                         \\
                                 & 70B                                  & \multicolumn{1}{c}{70.0*}               & 67.0                                       & \multicolumn{1}{c}{$66.0_{(-4.0)}$}                 & $\textbf{73.0}_{(+6.0)}$                                         \\ \hline
llama-3.1                        & 8B                                   & \multicolumn{1}{c}{\textbf{48.0}*}                 & 49.0                                         & \multicolumn{1}{c}{$32.0_{(-16.0)}$}                 & $36.0_{(-13.0)}$                                         \\ \hline
\multirow{2}{*}{llama-3.2}       & 1B                                   & \multicolumn{1}{c}{\textbf{4.0}}                  & 0.0                                          & \multicolumn{1}{c}{$3.0_{(-1.0)}$}                  & $0.0_{(+0.0)}$                                          \\ 
                                 & 3B                                   & \multicolumn{1}{c}{\textbf{25.0}}                 & 23.0                                         & \multicolumn{1}{c}{$18.0_{(-7.0)}$}                 & $16.0_{(-7.0)}$                                         \\ \hline
llama-3.3                        & 70B                                  & \multicolumn{1}{c}{76.0}                 & 78.0                                          & \multicolumn{1}{c}{$83.0_{(+7.0)}$}                 & $\textbf{88.0}_{(+10.0)}$                                         \\ \bottomrule
\end{tabular}
\end{table}

In Table~\ref{tab:my-table2}, we incorporated Prompt Style 2 to further analyze the performance of various LLM models on the SOMADHAN dataset, comparing standard prompting with Chain of Thoughts (CoT) prompting in both zero-shot and few-shot settings. The aim of this experiment was to determine whether Prompt Style 2 could improve accuracy, particularly for Llama series models and newer models from the Deepseek and Qwen series.

For the Deepseek-r1-distill-qwen model, the highest accuracy of 51\% was achieved with CoT and few-shot prompting, showing a significant improvement over the 44\% accuracy observed with standard few-shot prompting. However, it is important to note that all responses were in English, which deviates from the desired output in Bengali. Even with Chain of Thought and few-shot prompting, the response language format could not be improved. Similarly, the Deepseek-r1-distill-llama model achieved an impressive \textbf{66\%} accuracy with CoT and few-shot prompting, a 7\% increase over the standard few-shot performance of 60\%. Unlike the previous Deepseek model, this model's responses were in Bengali, aligning with the intended language output.

For the Qwen-2.5 model with 32 billion parameters, the highest accuracy of \textbf{71\%} was achieved using Chain of Thought with few-shot prompting. This represents a 3\% improvement compared to the 68\% accuracy observed with standard few-shot prompting. On the other hand, the Qwen-qwq model, despite incorporating Chain of Thought and few-shot prompting, could not generate responses in Bengali. All responses from this model were in English, indicating that the language adaptation was not achieved even with the use of Chain of Thought.

The Llama-3 model with 8 billion parameters did not achieve better results when Chain of Thought (CoT) prompting was introduced. The highest accuracy was achieved with standard prompting in the zero-shot setting, which reached 47\%. Instead of increasing accuracy, the use of Chain of Thought actually decreased the accuracy percentage. Furthermore, despite this accuracy, all responses were in English rather than Bengali, which deviates from the intended output. On the other hand, the Llama-3 model with 70 billion parameters performed better, achieving 73\% accuracy compared to 67\% with standard prompting, reflecting a 6\% improvement with the use of CoT. The smaller model (8B) again struggled with Chain of Thought prompting even with a change in Prompt style, possibly due to its limited capacity for processing more intricate tasks or adapting to Bengali. The larger Llama-3 (70B), however, demonstrated a noticeable improvement with the new prompt style. This suggests that parameter size and prompt play a crucial role in the effectiveness of Chain of Thought prompting, especially when dealing with multi-step reasoning and language adaptation. The Llama-3.1 (8B), Llama-3.2 (1B), and Llama-3.2 (3B) models also did not show any improvements with the introduction of Chain of Thought (CoT) prompting. In fact, their accuracy remained below expectations, and in some cases, even decreased. The Llama-3.1 (8B) model, for instance, still achieved its highest accuracy with standard zero-shot prompting (48\%), and CoT prompted a decrease in accuracy. Similarly, Llama-3.2 (1B) and Llama-3B also failed to demonstrate any notable improvements with and without CoT prompting, and their performance remained subpar in both zero-shot and few-shot settings. Even with the introduction of Prompt Style 2, Llama-3.3 (70B) continues to achieve the highest accuracy, reaching an impressive \textbf{88\%} on our proposed dataset. This performance was attained with the inclusion of Chain of Thought (CoT) and few-shot prompting, showing a remarkable 10\% improvement over the standard few-shot prompting accuracy of 78\%. This substantial increase highlights the significant benefits of applying CoT prompting to advanced models like Llama-3.3, demonstrating its effectiveness in handling complex reasoning tasks. The model's ability to achieve the highest accuracy, even with the new prompt style, further underscores the potential of CoT for enhancing model performance, particularly for larger, more capable models like Llama-3.3.

The Chain of Thought (CoT) prompting has proven to be a powerful approach in solving complex reasoning tasks, particularly in the domain of mathematical problems in Bengali. By breaking down intricate problems into smaller, manageable steps, CoT enables the models to better understand the logical structure of the task and produce more accurate solutions. In our experiments, CoT significantly enhanced the performance of larger models like GPT-4o and Llama-3.3, demonstrating its ability to handle multi-step reasoning. Moreover, the introduction of different prompt styles, such as Prompt Style 2, has further improved the efficacy of CoT by better aligning the model's responses with the intended language output. This highlights the importance of fine-tuning both the prompting techniques and the model parameters to optimize performance, particularly in multilingual settings like Bengali, where language-specific nuances can significantly impact the model's ability to reason and generate accurate answers. By strategically incorporating CoT with an appropriate prompt style, we can enhance the model's capacity to tackle more complex tasks with higher accuracy and language adaptation, thus improving overall performance in real-world applications.

\subsection{LoRA Fintune based results}

% \usepackage{multirow}
\begin{table}[h]
\centering
\caption{Ablation Study with three variations of LoRA finetuning on SOMADHAN dataset. All metrics are Accuracy (\%).}
\label{tab:my-tableL}
\begin{tabular}{l ccc}
\hline
\multirow{1}{*}{\textbf{Parameter}}                                    &  \multicolumn{1}{c}{\textit{\textbf{Baseline}}}                                                                                             & \multicolumn{1}{c}{\textit{\textbf{\begin{tabular}[c]{@{}c@{}}Higher Rank \\ with Dropout\end{tabular}}}} & \textit{\textbf{\begin{tabular}[c]{@{}c@{}}Memory Efficient\\ with Lower Batch Size\end{tabular}}} \\ \hline
\textbf{LoRA Parameters}                                                          & \multicolumn{3}{c}{}                                                                                                                                                                                                                                                                                                                                         \\ 
r                                                                      & \multicolumn{1}{c}{16}                                                                                                                     & \multicolumn{1}{c}{32}                                                                                    & 16                                                                                                 \\
lora alpha                                                             & \multicolumn{1}{c}{16}                                                                                                                     & \multicolumn{1}{c}{32}                                                                                    & 16                                                                                                 \\ 
lora dropout                                                           & \multicolumn{1}{c}{0}                                                                                                                      & \multicolumn{1}{c}{0.1}                                                                                   & 0.1                                                                                                \\ 
target modules                                                         & \multicolumn{1}{c}{\begin{tabular}[c]{@{}c@{}}\{q\_proj, k\_proj, v\_proj, \\ o\_proj, gate\_proj, \\ up\_proj, down\_proj\}\end{tabular}} & \multicolumn{1}{c}{\begin{tabular}[c]{@{}c@{}}\{q\_proj, k\_proj, \\ v\_proj, o\_proj\}\end{tabular}}     & \begin{tabular}[c]{@{}c@{}}\{q\_proj, v\_proj, \\ o\_proj\}\end{tabular}                           \\ 
\begin{tabular}[c]{@{}l@{}}use gradient\\ checkpointing\end{tabular}   & \multicolumn{1}{c}{unsloth}                                                                                                                & \multicolumn{1}{c}{True}                                                                                  & True                                                                                               \\ \hline
\textbf{Training Arguments}                                            & \multicolumn{3}{c}{}                                                                                                                                                                                                                                                                                                                                         \\ 
learning rate                                                          & \multicolumn{1}{c}{2e-4}                                                                                                                   & \multicolumn{1}{c}{1e-4}                                                                                  & 2e-4                                                                                               \\ 
\begin{tabular}[c]{@{}l@{}}per device train \\ batch size\end{tabular} & \multicolumn{1}{c}{2}                                                                                                                      & \multicolumn{1}{c}{4}                                                                                     & 1                                                                                                  \\ 
\begin{tabular}[c]{@{}l@{}}gradient accumulation\\ steps\end{tabular}  & \multicolumn{1}{c}{4}                                                                                                                      & \multicolumn{1}{c}{2}                                                                                     & 8                                                                                                  \\ 
warmup steps                                                           & \multicolumn{1}{c}{5}                                                                                                                      & \multicolumn{1}{c}{10}                                                                                    & 5                                                                                                  \\ 
max steps                                                              & \multicolumn{1}{c}{60}                                                                                                                     & \multicolumn{1}{c}{100}                                                                                   & 60                                                                                                 \\ \hline
 
\textbf{Performance}                                                               & \multicolumn{1}{c}{\textbf{13.0}}                                                                                                                   & \multicolumn{1}{c}{\textbf{12.0}}                                                                                  & \textbf{17.0}                                                                                               \\ \hline
\end{tabular}
\end{table}

From Table~\ref{tab:my-tableL}, \textbf{Baseline Configuration (Finetune 1):} This setup, characterized by a rank of 16, no dropout, and a moderate learning rate of 2e-4, achieved an accuracy of 13\%. The simplicity of this configuration likely facilitated quicker optimization but also increased the risk of overfitting due to the lack of regularization. The moderate parameters provided a reference point for evaluating more complex setups, showing the limitations of basic fine-tuning without dropout.

\textbf{Higher Rank with Dropout (Finetune 2):} Incorporating a higher rank (32) and dropout (0.1), this configuration aimed to increase model capacity and regularization. However, it achieved a slightly lower accuracy of 12\%, despite its cautious learning rate of 1e-4. This result suggests that the added complexity and regularization may not align well with the dataset’s requirements, possibly due to over-parameterization or insufficient data to fully utilize the additional capacity.

\textbf{Memory-Efficient with Lower Batch Size (Finetune 3):} The third configuration utilized memory-efficient adjustments, including a lower batch size and increased gradient accumulation steps, alongside a higher learning rate of 2e-4. This setup achieved the highest accuracy of 17\%. The reduced batch size likely allowed for better gradient updates, while the increased accumulation steps improved learning stability. The higher learning rate might have further facilitated effective convergence by optimizing balancing and learning dynamics.

\subsection{Gpt-3.5 Finetune result}

% \usepackage{multirow}
\begin{table}[h]
\centering
\caption{Performance comparison of finetuned Gpt-3.5 with Gpt-4o. All metrics are Accuracy (\%).}
\label{tab:my-tableF}
\begin{tabular}{c c c c}
\hline
\textbf{Models}                                                                    & \textbf{Learning Rate} & \textbf{Epochs} & \textbf{Performance} \\ \hline
\multirow{3}{*}{Gpt-3.5 turbo-0125}                                                & \multirow{3}{*}{0.1}   & 5               & 12.0                 \\  
                                                                                   &                        & 10              & 23.0                 \\  
                                                                                   &                        & 15              & 12.0                 \\ \hline
\begin{tabular}[c]{@{}c@{}}Gpt-4o \\ (Standard Prompting\\ Zero Shot)\end{tabular} & -                      & -               & 79.0                 \\ \hline
\end{tabular}
\end{table}

Table~\ref{tab:my-tableF} presents a performance comparison between the fine-tuned GPT-3.5 (turbo-0125) and GPT-4o, evaluated using standard zero-shot prompting. The results reveal notable differences in accuracy. To build the fine-tuned model, we utilized 50 examples with appropriate instructions, including explicit chain-of-thought reasoning for complex problems. This approach ensured that the model not only provided answers but also demonstrated the reasoning process behind solving math problems. The examples followed the JSONL file format as outlined in the OpenAI API documentation. Figure~\ref{fig:json} illustrates an example of how the JSON file was structured to create the fine-tuned model. Fine-tuning GPT-3.5 with a learning rate of 0.1 over 10 epochs achieved the best performance, with an accuracy of 23.4\%. However, increasing the number of epochs from 10 to 15 did not yield any further improvements, as the model's performance plateaued at 12.6\%, which was equivalent to the performance achieved at 5 epochs. This suggests that fine-tuning for a higher number of epochs may lead to diminishing returns or potential overfitting.

In contrast, GPT-4o, which was not fine-tuned but instead evaluated using its standard zero-shot prompting capability, demonstrated a significantly superior performance of 79.2\%. This highlights GPT-4o’s inherent ability to generalize across tasks without the need for task-specific fine-tuning, significantly outperforming GPT-3.5, even after fine-tuning.

Appendix~\ref{Appendix-A} illustrates the OpenAI Playground examples and their outputs and presents the responses of Llama 3.3 on Bengali Math Word Problems, and Appendix~\ref{Appendix-B} showcases the correct and incorrect responses of GPT models on Bengali Math Word Problems.

% Table~\ref{tab:my-tableF} shows the performance comparison between fine-tuned GPT-3.5 (turbo-0125) and GPT-4.0 (using standard zero-shot prompting), reveals significant differences in accuracy. Fine-tuning GPT-3.5 with a learning rate of 0.1 over 10 epochs achieved the best performance among the fine-tuned configurations at 23.4\%. However, increasing the epochs to 15 did not yield any further improvements, with the performance plateauing at 12.6\%, matching the result from 5 epochs. This suggests that fine-tuning at higher epochs might lead to overfitting or diminished returns.

% In contrast, GPT-4.0, without any fine-tuning, demonstrated a far superior performance of 79.2\% using its standard zero-shot prompting capability. This highlights GPT-4.0's inherent ability to generalize and perform well on tasks without the need for task-specific training, significantly outperforming GPT-3.5 even with fine-tuning

\section{Conclusion and Future Work}
In this study, we presented a novel approach for solving Bengali Math Word Problems (MWPs) using Large Language Models (LLMs) with Chain of Thought (CoT) prompting. We introduced the SOMADHAN dataset, specifically designed to address Bengali MWPs requiring complex reasoning and multi-step solutions. Our experiments demonstrated that CoT prompting significantly improved the performance of LLMs, particularly for reasoning-intensive tasks in Bengali, a low-resource language. Among the models evaluated, Llama-3.3 outperformed all other models, achieving the highest accuracy of 88\% with CoT and few-shot prompting. The GPT-4o model, with its substantial number of parameters, also performed exceptionally well, showing high accuracy in both zero-shot and few-shot settings. Additionally, we leveraged Low-Rank Adaptation (LoRA) fine-tuning techniques to optimize large models for Bengali MWPs. Our findings contribute to advancing the field of Bengali language processing and educational technologies, providing a foundation for future work in multilingual problem-solving tasks.

While our approach showed promising results, several limitations need to be acknowledged. First, due to resource constraints and token costs, we limited our evaluation to 1000 testing samples from both the SOMADHAN and PatiGonit datasets. This reduced sample size might have impacted the generalizability of the results, and future studies could benefit from evaluating a larger subset of the dataset. Additionally, while CoT prompting improved model performance in many cases, the language adaptation for smaller models, particularly the LLaMA series, was not optimal. Models such as Llama-3 (8B) struggled to effectively solve Bengali MWPs, indicating that smaller models may not perform as well as larger ones when it comes to complex reasoning tasks in low-resource languages. The reliance on English responses for some models, even after incorporating CoT prompting, also highlights the challenge of multilingual task adaptation.

For future research, expanding the SOMADHAN dataset to include a larger variety of Bengali Math Word Problems with more diverse reasoning steps could improve model training and evaluation. Future work should also focus on the development of even more efficient fine-tuning strategies, such as further optimizations of Low-Rank Adaptation (LoRA), to handle the increasing complexity of language models without incurring high computational costs. Investigating methods to ensure better language output alignment in multilingual models, such as fine-tuning for Bengali language generation, would be an essential step toward improving the overall effectiveness of LLMs in real-world applications.

\subsubsection*{Author Contributions}
\textbf{Bidyarthi Paul: }Methodology, Writing-Original draft, Visualization, Validation \\
\textbf{Jalisha Jashim Era: }Methodology, Investigation, Writing-Original draft, Conceptualization \\
\textbf{Mirazur Rahman Zim: }Data Curation, Resources, Visualization
\\
\textbf{Tahmid Sattar Aothoi: }Data Curation, Software, Visualization
\\
\textbf{Mr. Faisal Muhammad Shah: }Supervision, Writing-review\& editing

\subsubsection*{Acknowledgments}
This research did not receive any specific grant from funding agencies in the public, commercial, or not-for-profit sectors.

\bibliographystyle{unsrt}  
\bibliography{references}  %%% Remove comment to use the external .bib file (using bibtex).


% George Kour and Raid Saabne.
% \newblock Real-time segmentation of on-line handwritten arabic script.
% \newblock In {\em Frontiers in Handwriting Recognition (ICFHR), 2014 14th
%   International Conference on}, pages 417--422. IEEE, 2014.

% George Kour and Raid Saabne.
% \newblock Fast classification of handwritten on-line arabic characters.
% \newblock In {\em Soft Computing and Pattern Recognition (SoCPaR), 2014 6th
%   International Conference of}, pages 312--318. IEEE, 2014.

% \bibitem{hadash2018estimate}
% Guy Hadash, Einat Kermany, Boaz Carmeli, Ofer Lavi, George Kour, and Alon
% \newblock Estimate and replace: A novel approach to integrating deep neural
%   networks with existing applications.
% \newblock {\em arXiv preprint arXiv:1804.09028}, 2018.

% \end{thebibliography}

% Start a new page
\clearpage

\appendix

\section{Prompting in OpenAI Playground and their Output examples along with Response of Llama 3.3 on Bengali Math Word Problems}
\label{Appendix-A}

\begin{figure}[H]
    \centering
    \includegraphics[width=1.0\linewidth]{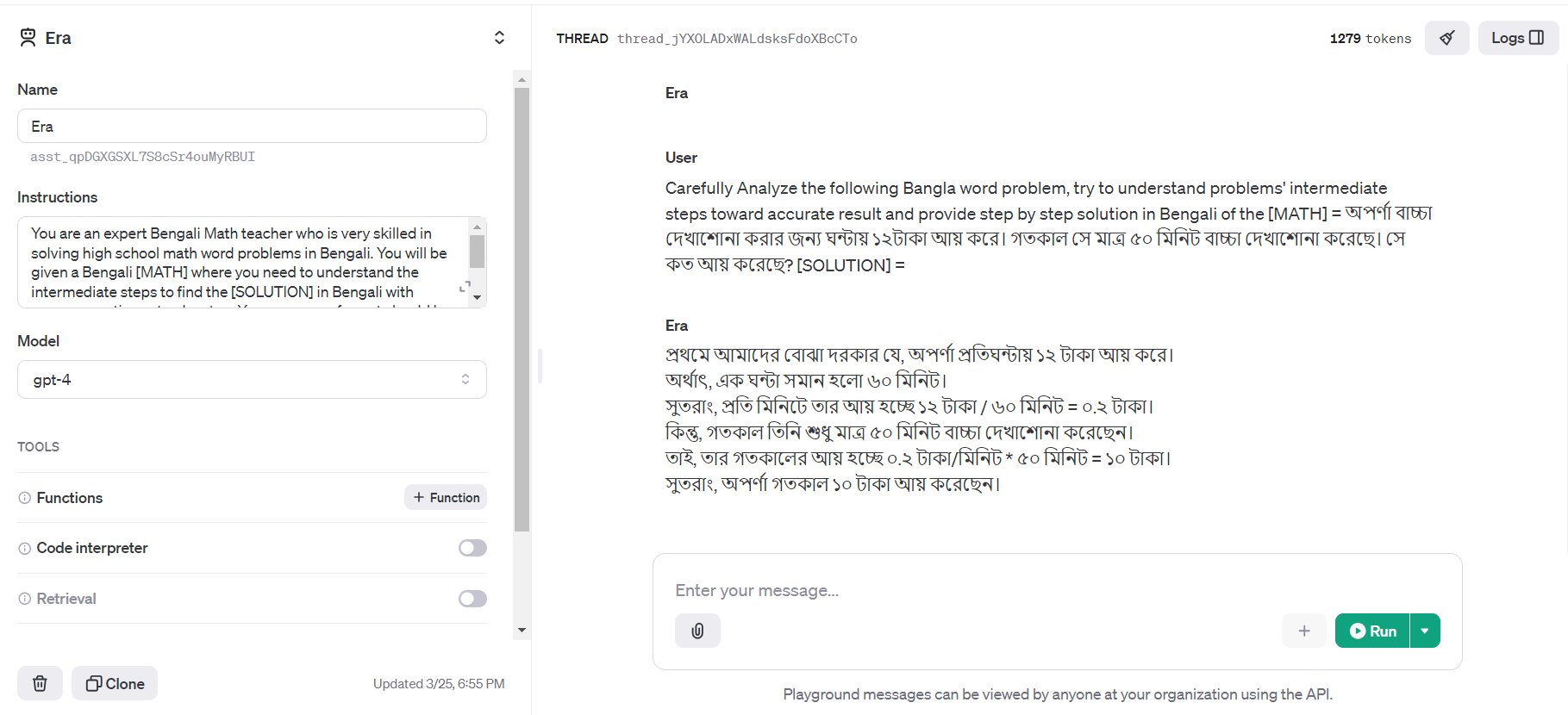}
    \caption{SOMADHAN dataset, Prompting and Output in OpenAI Playground}
    \label{fig:llm1}
\end{figure}

\begin{figure}[H]
    \centering
    \includegraphics[width=1.0\linewidth]{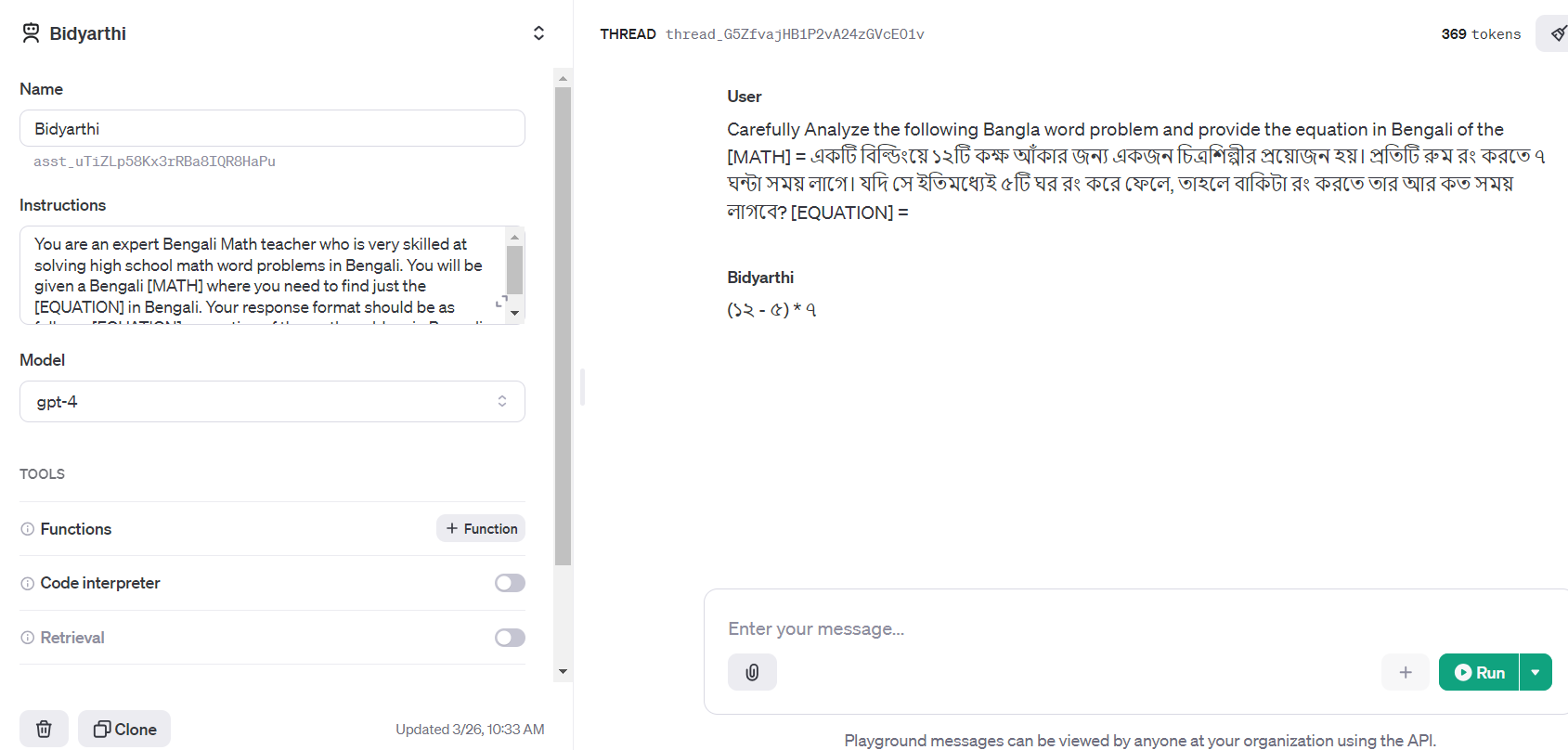}
    \caption{PatiGonit dataset, Prompting and Output in OpenAI Playground}
    \label{fig:llm2}
\end{figure}

% \begin{figure}[!htbp]
%     \centering
%     \includegraphics[width=1.0\linewidth]{Llama response.png}
%     \caption{Response generated by LLaMA 3.3 model on a Bengali Math Word Problem using Chain-of-Thought reasoning.}
%     \label{fig:llama3.3-response}
% \end{figure}

% \section{Response of Llama 3.3 on Bengali Math Word Problems}
% \label{Appendix-B}

\begin{figure}[H]
    \centering
    \includegraphics[width=\linewidth]{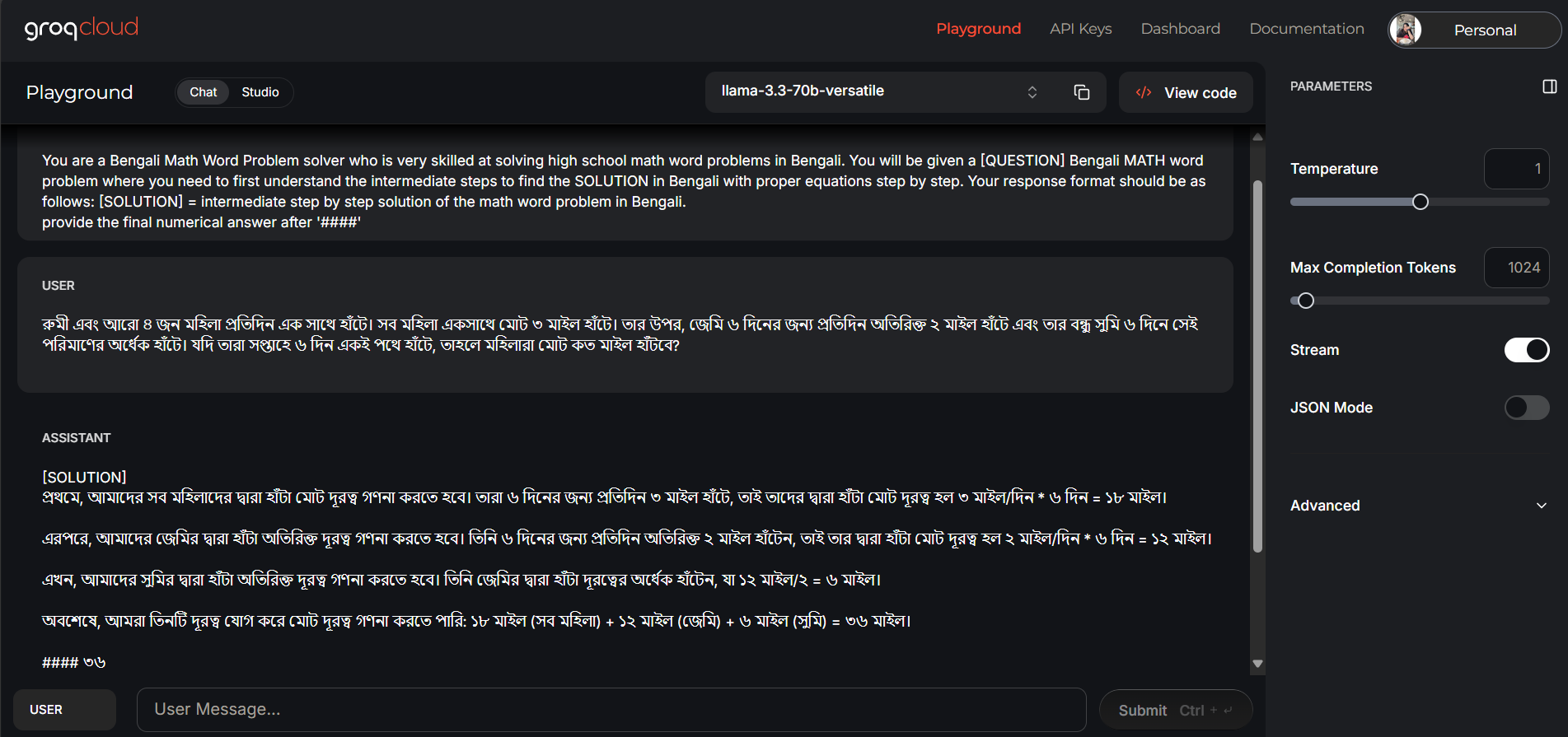}
    \caption{Response generated by LLaMA 3.3 model on a Bengali Math Word Problem using Chain-of-Thought reasoning.}
    \label{fig:llama3.3-response}
\end{figure}

\section{Correct and Incorrect Responses of GPT Models on Bengali Math Word Problems}
\label{Appendix-B}

\begin{figure}[h]
    \centering
    \includegraphics[width=\linewidth]{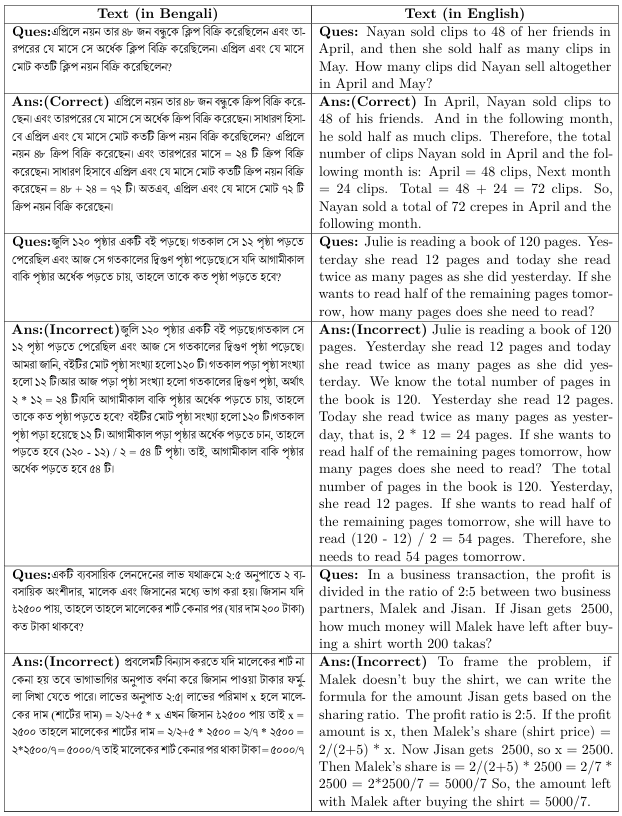}
    \caption{Example of correct chains of thought produced by the GPT-3.5 for the SOMADHAN dataset for Few Shot}
    \label{fig:example1}
\end{figure}

\begin{figure}[h]
    \centering
    \includegraphics[width=\linewidth]{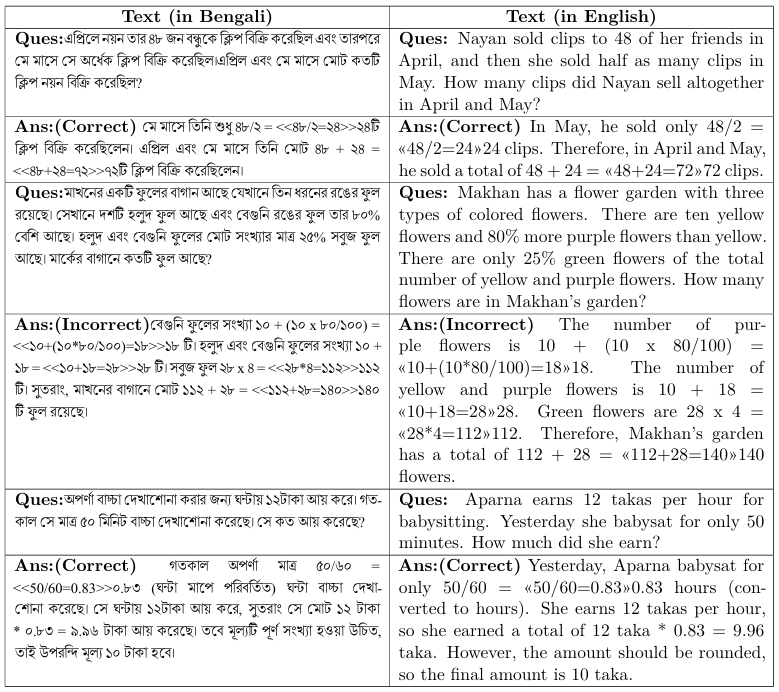}
    \caption{Example of correct chains of thought produced by the GPT-4o for the SOMADHAN dataset for Few Shot}
    \label{fig:example2}
\end{figure}

\begin{figure}[h]
    \centering
    \includegraphics[width=\linewidth]{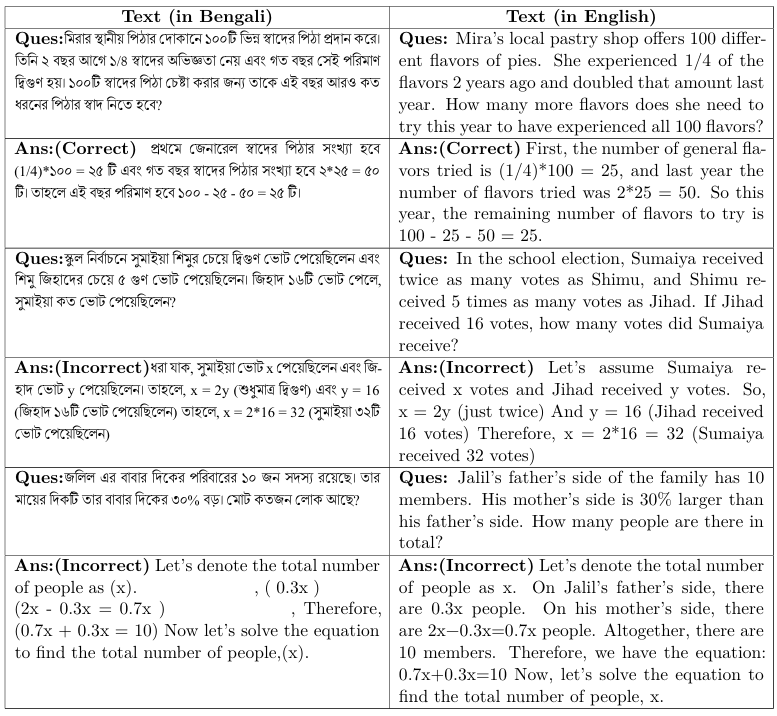}
    \caption{Example of correct chains of thought produced by the GPT-3.5 for the SOMADHAN dataset for Fine Tuning}
    \label{fig:example3}
\end{figure}

\end{document}